%% file: acl_latex.tex
\pdfoutput=1

\documentclass[11pt]{article}

\usepackage[final]{acl}

\usepackage{times}
\usepackage{latexsym}
\usepackage{xcolor}
\usepackage{colortbl}
\usepackage[T1]{fontenc}
\usepackage[utf8]{inputenc}

\usepackage{microtype}
\usepackage{inconsolata}

\usepackage{amsfonts}
\usepackage{amssymb}
\usepackage{pifont}
\usepackage{tabularx}
\usepackage{arydshln}
\usepackage{mathtools}
\usepackage{adjustbox}
\usepackage{multirow}
\usepackage{paralist} 
\usepackage{CJKutf8} 
\usepackage{booktabs} 

\usepackage{nicefrac}
\usepackage{custom} 

\usepackage{enumerate}
\usepackage{bbding} 
\usepackage{wasysym}

\usepackage{caption}
\usepackage{subcaption}

\usepackage[ruled,linesnumbered]{algorithm2e}
\SetKwInOut{Parameter}{Parameters}

\usepackage{amsthm} 
\theoremstyle{definition}

\usepackage{tikz}
\usepackage[edges]{forest} 
\usepackage{arydshln} 

\usepackage{pgfplots}
\usepackage{makecell}

\usepackage{dsfont}
\usepackage{courier}
\usepackage[most]{tcolorbox}
\definecolor{blue1}{rgb}{0.15,0.15,0.15}
\definecolor{blue2}{rgb}{0.30,0.30,0.30}
\definecolor{blue3}{rgb}{0.45,0.45,0.45}
\definecolor{blue4}{rgb}{0.60,0.60,0.60}
\definecolor{blue5}{rgb}{0.70,0.70,0.70}
\definecolor{blue6}{rgb}{0.80,0.80,0.80}
\newtcolorbox{taskbox}[2][]{
	enhanced, breakable,
	colframe=blue3!40,
	colback=blue5!5,
	arc=1mm,
	outer arc=1mm,
	fontupper=\small,
	fontlower=\small,
	coltitle=blue1,
	fonttitle=\bfseries,
	boxsep=1mm,
	left=0mm,
	right=0mm,
	top=0mm,
	bottom=0mm,
	before={\noindent},
	segmentation style={solid, blue3},
	title=#2,
	#1
}

\definecolor{tkcolor}{RGB}{224,223,255}
\newtcolorbox{takeaways}[2][]{
	width=\columnwidth,
	colback = tkcolor, 
	colframe = tkcolor, 
	boxsep=0pt,left=10pt,right=10pt,top=2pt,bottom=3pt,
	fontupper=\linespread{0.9}\selectfont,
	title=#2,#1}

\newtcolorbox{mybox}[2][]{
	width=\columnwidth,
	colback = gray!8, 
	colframe = gray!8, 
	boxsep=0pt,left=10pt,right=10pt,top=0pt,bottom=0pt,
	fontupper=\linespread{0.9}\selectfont,
	title=#2,#1}
\newcommand{\datasetname}{\texttt{M$^3$CoT}}

\definecolor{darkolivegreen}{rgb}{0.33, 0.42, 0.18}
\definecolor{my_green}{RGB}{40,154,121}
\definecolor{my_yellow}{RGB}{255,165,0}
\definecolor{my_red}{RGB}{176,46,46}

\usepackage{hyperref}

\usepackage{pifont}       
\usepackage{bbding}       
\usepackage{fontawesome}  

\newcommand{\correctmark}{\textcolor{my_green}{\ding{52}}} 
\newcommand{\errormark}{\textcolor{my_red}{\ding{56}}}

\title{M$^3$CoT: A Novel Benchmark for Multi-Domain Multi-step Multi-modal Chain-of-Thought}

\author{Qiguang Chen$^{\spadesuit}$ \quad Libo Qin$^{\clubsuit}$\thanks{\ \ Corresponding Author} \quad Jin Zhang$^{\spadesuit}$ \quad Zhi Chen$^{\diamondsuit}$ \\
	\textbf{Xiao Xu$^{\spadesuit}$ \quad Wanxiang Che$^{\spadesuit}$}\footnotemark[1] \\
	$^{\spadesuit}$ Research Center for Social Computing and Information Retrieval \\
	$^{\spadesuit}$ Harbin Institute of Technology, China\\
	$^{\clubsuit}$  School of Computer Science and Engineering, Central South University, China \\
	$^{\diamondsuit}$ Shanghai AI  Laboratory\\
	\texttt{\{qgchen,car\}@ir.hit.edu.cn}, \texttt{lbqin@csu.edu.cn}\\}

\begin{document}
\maketitle
\input{sections/abstract}

\input{sections/introduction}

\input{sections/approach}

\input{sections/experiment}
\input{sections/related}
\input{sections/conclusion}

\bibliography{custom}
\bibliographystyle{acl}
\clearpage
\appendix

\input{sections/appendix}

\end{document}

%% file: sections/abstract.tex
\begin{abstract}
Multi-modal Chain-of-Thought (MCoT) requires models to leverage knowledge from both textual and visual modalities for step-by-step reasoning, which gains increasing attention. 
	Nevertheless, the current MCoT benchmark still faces some challenges: (1) \textit{absence of visual modal reasoning}, (2) \textit{single-step visual modal reasoning}, and (3) \textit{Domain missing}, thereby hindering the development of MCoT.	 
Motivated by this, we introduce a novel benchmark (\datasetname{}) to address the above challenges, advancing the multi-domain, multi-step, and multi-modal CoT.
Additionally, we conduct a thorough evaluation involving abundant MCoT approaches on Vision Large Language Models (VLLMs). 
In addition, we highlight that the current VLLMs still struggle to correctly reason in \datasetname{} and there remains a large gap between existing VLLMs and human performance in \datasetname{}, despite their superior results on previous MCoT benchmarks. 
To our knowledge, we take the first meaningful step toward the multi-domain, multi-step, and multi-modal scenario in MCoT.
We hope that \datasetname{} can serve as a valuable
resource, providing a pioneering foundation in multi-domain, multi-step, multi-modal chain-of-thought research.
\end{abstract}

%% file: sections/introduction.tex
\section{Introduction}

\begin{figure}[t]
	\centering
	\includegraphics[width=0.49\textwidth]{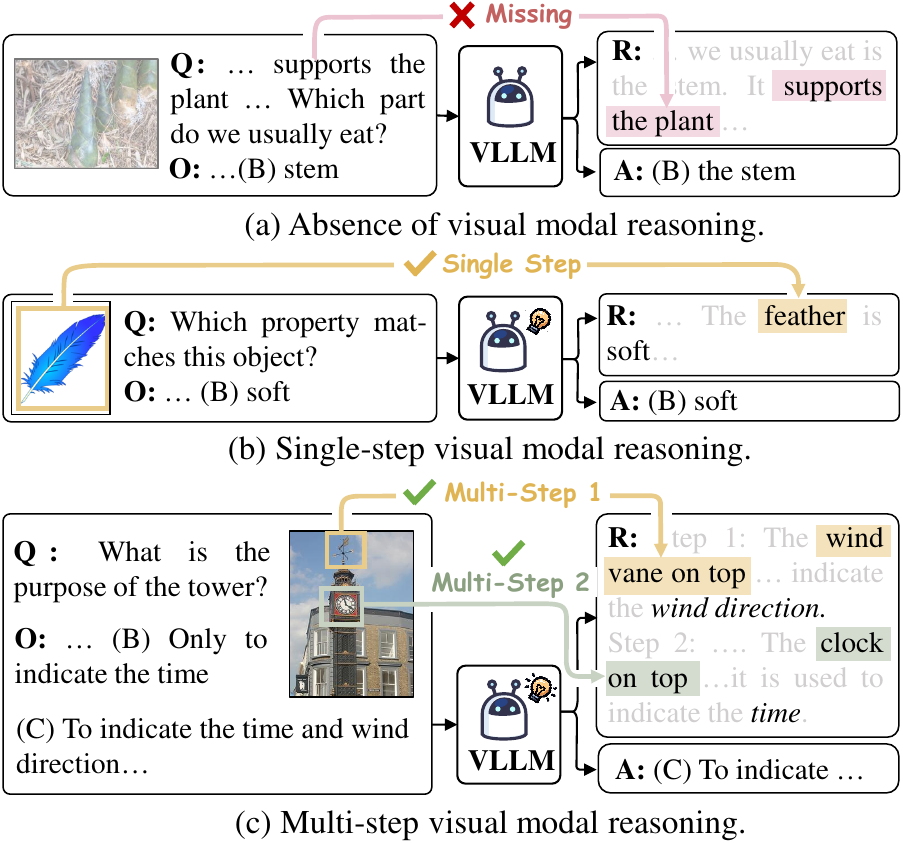}
	\caption{
				The example of Absence of visual modal reasoning (a), Single-step visual modal reasoning (b), and Multi-step visual modal reasoning (c). 
				\textbf{Q}: textual question;  \textbf{O}: textual options; \textbf{R}: generated rationale; \textbf{A}: generated answer.
	}
	\label{fig:instro}
\end{figure}

Recent advancements in Large Language Models (LLMs) have led to notable improvements in Chain-of-Thought (CoT) in textual modality~\cite{wei2022emergent,wei2022chain,wang-etal-2023-plan,hu2024treeplanner}.
In addition, some works begin to extend the textual CoT capabilities to multi-modal CoT reasoning (MCoT). 
Take Figure~\ref{fig:instro} (c) as an example, 
multi-modal CoT requires both the visual and textual features to generate a rationale and a final answer.
To this end, \citet{lu2022learn} introduced ScienceQA benchmark and laid the foundation for MCoT.
Subsequently, \citet{zhang2023multimodal} proposed
a two-stage approach during multi-modal reasoning for MCoT.
Additionally, \citet{wang2023t} developed T-SciQA framework to distill high-quality rationales from ChatGPT,
which attains an average accuracy of 96.2\%, surpassing even  the human accuracy of 88.4\%.

\begin{figure*}[t]
	\centering 
	\subfloat[]{
		\label{fig:subfig1}\includegraphics[width=0.27\textwidth]{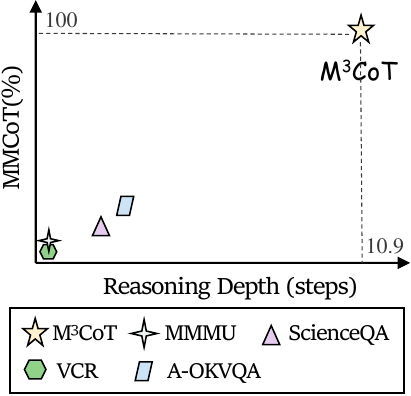}
	}
	\subfloat[]{
		\begin{adjustbox}{width=0.70\textwidth}
			\begin{tabular}{{l}*{8}{c}}
				\toprule	
				& \multirow{2}{*}{\textbf{\# Q}} & \multirow{2}{*}{\textbf{\# I}} & \multicolumn{3}{c}{\textbf{Domain}} & \multirow{2}{*}{\textbf{MMCoT}} & \multirow{2}{*}{\textbf{Rationale}}  \\ 
				\cmidrule{4-6}
				&  &&\textbf{Science} & \textbf{Mathmatic} & \textbf{Commonsense} & & \\ 
				\midrule
				\text{Geometry3K} \cite{lu2021inter} & 3,002 &2,342 
								 & \errormark & \correctmark & \errormark & \errormark & \errormark \\
				\text{TQA}~\cite{kembhavi2017you} & \textbf{26,260} & 3,455 
								& \errormark & \errormark & \correctmark & \errormark & \errormark \\
				MathVista~\cite{lu2023mathvista} & 5,487 & 6,141
								 & \errormark & \correctmark & \errormark & \errormark & \errormark  \\
				MME~\cite{fu2023mme} & 2,194 & 1,097
								 & \errormark & \errormark & \correctmark & \errormark & \errormark \\
				SeedBench~\cite{li2023seed} & - & 19,242 
								& \errormark & \errormark & \correctmark & \errormark & \errormark \\
				MM-Vet~\cite{yu2023mm} & 205 & 187
								 & \errormark & \correctmark & \correctmark & \errormark & \errormark \\
				\midrule
				
				\text{VCR}~\cite{zellers2019recognition}  & 290k & 99,904 & \errormark & \errormark & \correctmark & $\sim$4\% & \correctmark \\
				\text{A-OKVQA}~\cite{schwenk2022okvqa} & 24,903 &  23,692 & \errormark & \errormark & \correctmark & $\sim$21\% & \correctmark \\
				\text{KI-VQA}~\cite{li2023comprehensive} & 4,290 &  4,189
								 & \errormark & \errormark & \correctmark & $\sim$17\% & \correctmark \\
				ScienceQA~\cite{lu2022learn} & 21,208 & 10,332 
								 & \correctmark & \errormark & \errormark & $\sim$8\% & \correctmark \\
				\multirow{2}{*}{MMMU~\cite{yue2023mmmu}} & \multirow{2}{*}{11,550} & \multirow{2}{*}{11,264} & \multirow{2}{*}{\correctmark} & \multirow{2}{*}{\correctmark} & \multirow{2}{*}{\errormark} & \multirow{2}{*}{$\sim$8\%} & Science Only \\
				&&&&&&& (<18\%) \\
				\midrule
				\datasetname{} (ours) & 11,459 &  11,293 
								 & \correctmark & \correctmark & \correctmark & \textbf{100\%} & \correctmark \\
				\bottomrule
			\end{tabular}
		\end{adjustbox}
	}
	\caption{
				Comparison of \datasetname{} and multi-modal related datasets on (a) MCoT reasoning complexity and (b) detailed diversity. MMCoT: the ratio of samples with  multi-step MCoT (MMCoT) in the datasets; \textbf{\#X}: the size of X, \textbf{Q}: Question; \textbf{I}: Image. 
				The simplicity of the previous benchmarks lies in its MMCoT, domain, and reasoning depth. We will describe the details of the corresponding statistics in Appendix~\ref{append:exist-data}.}
	\label{exp:data_comparison}
\end{figure*}

Inspired by the recent remarkable advancements in the MCoT literature (surpassing human performance), we seek to explore an interesting question: \textit{Has MCoT task been solved perfectly?} In our deep analysis, the conclusion is definitely ``\textit{NO}''.
As shown in Figure~\ref{exp:data_comparison} (a), we observe that current benchmarks are too simple, leading to an overestimation of current progress. 
Furthermore, we find that the existing benchmarks exhibit three major drawbacks (see Figure~\ref{exp:data_comparison} (b)):
(1) \textbf{\textit{Absence of visual modal reasoning}}: 
As shown in Figure~\ref{fig:instro} (a), the model can successfully produce rationale and answer solely based on the textual modality context of ``\textit{supports the plant}'', which cannot truly reflect the ability of multi-modal CoT model.
(2) \textbf{\textit{Single-step visual modal reasoning}}:
As illustrated in Figure~\ref{fig:instro} (b), the model only requires a single-step ``feather'' object to predict the correct rationale and answer, which cannot be satisfied in the complex multi-step CoT scenario.
(3) \textbf{\textit{Domain Missing}}:
Commonsense and mathematics are important domains for evaluating multi-modal CoT~\citep{wei2022chain,qin2023cross}, but the current benchmarks lack these topics, 
hindering the comprehensive evaluation progress of multi-modal CoT.
Nevertheless, in real-world scenarios, multi-step MCoT reasoning is frequently observed in diverse domains. For example, as illustrated in Figure~\ref{fig:instro} (c), vision large language models (VLLMs) are required to identify the correct options integrating at least two multi-modal reasoning steps (as indicated by the orange and green lines.). Such multi-step MCoT tasks are required to effectively perform multi-step reasoning across multiple modalities, which cannot be achieved by previous single-step multi-modal CoT approaches.

Motivated by these observations and issues, we introduce a novel benchmark about multi-domain multi-step multi-modal chain-of-thought reasoning (\datasetname{}) based on ScienceQA~\cite{lu-2023-science}.
Specifically, to address the first issue, we directly remove samples that could infer the final answer without the need for images. To tackle the second issue, we manually annotate and select multi-step multi-modal samples. Specifically, we provide expert annotators with textual context and rationales without images. Experts are required to determine when multi-step reasoning cannot be resolved solely based on textual context. Subsequently, we present the images to experts to ascertain whether multi-step reasoning occurred across textual and visual modalities. To solve the third issue, we explore LLM-guided augmentation to synthesize the multi-step MCoT data for commonsense and mathematics domains.
We evaluate abundant representative MCoT approaches on \datasetname{} in extensive scenarios, yielding several {\textbf{key takeaways}}: (1) \textit{VLLM shows CoT emergence phenomenon  at the parameter level over 10 billion} ($\ge$ 13B); (2) \textit{Fine-tuning has better hope on multi-step MCoT, compared with the failures of vanilla in-context-learning, tool usage, and prompting strategies}. (3) \textit{\datasetname{} is tough enough and all methods still struggle compared with human performance}.

In conclusion, the primary contributions of our work are summarized as follows:
\begin{figure*}[t]
	\centering
	\includegraphics[width=0.99\textwidth]{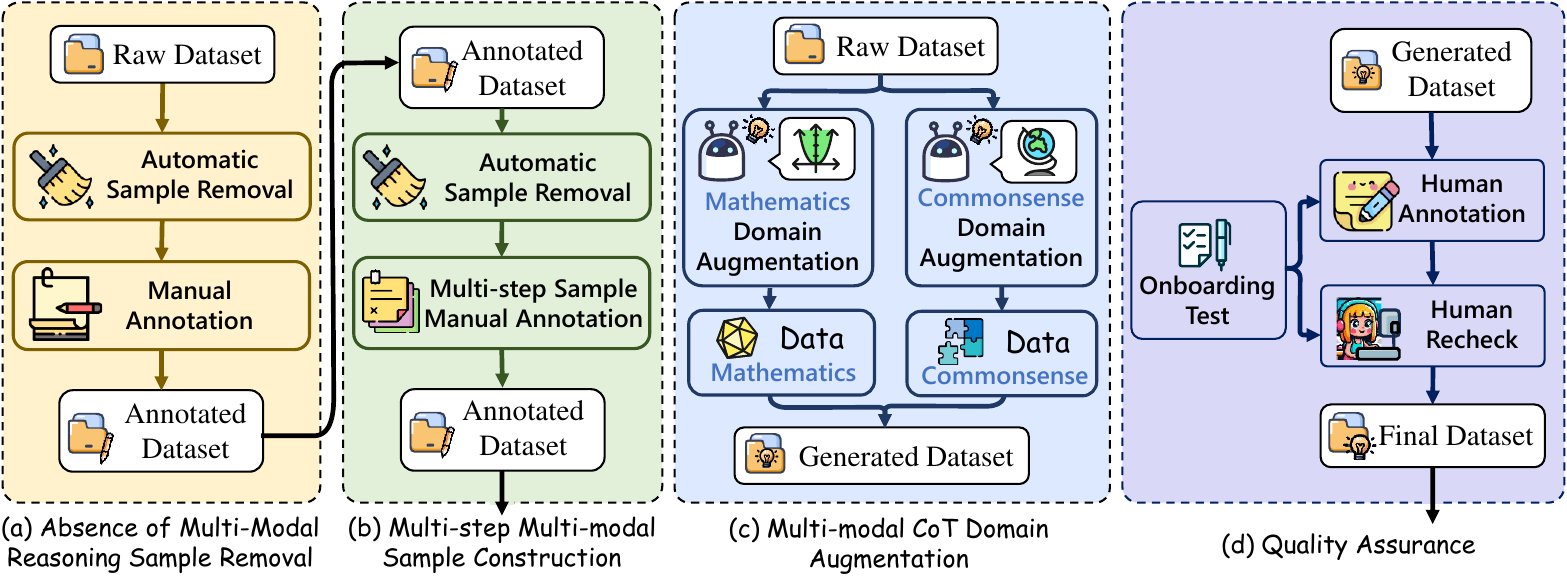}
	\caption{
	Dataset construction workflow including (a) \textit{Absence of Visual Modal Reasoning Sample Removal} ($\S$~\ref{sec:clean}), (b) \textit{Multi-step Multi-modal Sample Construction} ($\S$~\ref{sec:multi-step}), (c) \textit{Multi-modal CoT Domain Augmentation} ($\S$~\ref{sec:data-generation}), and (d) \textit{Quality Assurance} ($\S$~\ref{sec:quality}).
	}
	\label{fig:data_annotation}
\end{figure*}
\begin{itemize}
	\setlength{\itemsep}{0.2em}
	\item We identify the weaknesses of current multi-modal CoT benchmarks that can not handle complex multi-step reasoning scenarios, which motivates researchers to rethink the current progress of multi-modal CoT.
	
	\item To the best of our knowledge, we are the first to consider the multi-domain, multi-step, multi-modal CoT scenario and introduce \datasetname{} to this end.
	
	\item We evaluate abundant representative MCoT approaches on \datasetname{} and summarize some insightful takeaways, hoping to inspire more breakthroughs in this direction.
	
\end{itemize}

To facilitate further research, all data and code are available at \url{https://github.com/LightChen233/M3CoT}.

\section{Problem Formalization}
\label{sec:preliminary}
This section describes the definition of multi-step multi-modal CoT.
Specifically, unlike the traditional textual CoT, multi-step multi-modal CoT should consider a scenario involving an image $I$, a question $Q$, a context $C$ and a set of $n$ options $\mathcal{O}= \{o_1, ..., o_n\}$. First we construct a textual prompt $\mathcal{T}$:
\begin{equation}
	\mathcal{T} = \texttt{Prompt}(Q, C, \mathcal{O}),
\end{equation}
where $\texttt{Prompt}(\cdot)$ represents any method used to convert textual inputs into an instruction format.

Then, model should generate a step-wise rationale $\mathcal{R}_m=\{S_1, ..., S_m\}$, each step determined by\footnote{All step segmentation in this paper follows the ROSCOE~\cite{golovneva2023roscoe}.}:
\begin{equation}
	S_i = \underset{S_i \in \mathcal{R}_m}{\operatorname{argmax}} \ P(S_i|I, \mathcal{T}),
\end{equation}
\begin{equation}
	P(S_i|I, \mathcal{T})\! =\! \begin{cases}
		P(S_i|\mathcal{T}, \mathcal{R}_{i-1}),\! & S_i \notin \mathcal{S}; \\
		P(S_i|I, \mathcal{T}, \mathcal{R}_{i-1}),\! & S_i \in \mathcal{S},
	\end{cases}
\end{equation}
where $\mathcal{S}$ indicates steps that require multi-modal reasoning. This reasoning is considered multi-step and multi-modal if $|\mathcal{S}|\ge 2$.

Finally, the model arrives at the final answer $\mathcal{Y}$, which is denoted as:
\begin{equation}
	\mathcal{Y} = \underset{o \in \mathcal{O}}{\operatorname{argmax}} \ P(o|\mathcal{R}_m).
\end{equation}

%% file: sections/approach.tex
\section{Dataset Annotation}

This section describes the 
annotation process of \datasetname{}, including: \textit{Absence of Visual Modal Reasoning Sample Removal} ($\S$~\ref{sec:clean}), \textit{Multi-step MCoT Sample Construction} ($\S$~\ref{sec:multi-step}), \textit{MCoT Domain Augmentation} ($\S$~\ref{sec:data-generation}), and \textit{Quality Assurance} ($\S$~\ref{sec:quality}). The samples we generate and retain at each stage are detailed in Figure~\ref{fig:annotaion}.

\subsection{Absence of Visual Modal Reasoning Sample Removal}
\label{sec:clean}
This section focuses on addressing the absence of visual modal reasoning challenge from ScienceQA.

\noindent\textbf{\textit{Automatic Sample Removal}: } First, we directly filter out samples without images,  thereby refining the dataset to include only those samples that potentially require multi-modal reasoning.

\noindent\textbf{\textit{Manual Annotation}: }Despite the automatic process, some samples containing images are still irrelevant for multi-modal reasoning (see Figure~\ref{fig:instro} (a)). Therefore, we further employ manual annotation, requiring experts to verify whether each sample meets the criteria for MCoT. Specifically, our annotation process and instructions are shown in Appendix~\ref{sec:annotate-first}.

\subsection{Multi-step MCoT Sample Construction}
\label{sec:multi-step}
This section aims to incorporate multi-step reasoning characteristics from the last processed data.

\noindent\textbf{\textit{Automatic Sample Removal: }}
In this step, we first automatically filter out some simple samples with overly simplistic rationales, which comprise less than two steps. 
By doing this, we can reduce the manual annotation burden and increase the reliability of \datasetname{}. More details are illustrated in Appendix~\ref{sec:annotate-second}.

\noindent\textbf{\textit{Multi-step Sample Manual Annotation: }}
After automatic sample removal, we further utilize manual annotation to obtain the final multi-step multi-modal reasoning dataset.
Specifically, 
human experts are first provided with textual context and rationales without visual modality. They are focused on determining whether it is necessary to answer the samples multiple times based on the visual content. Once experts find that multi-step reasoning needs multiple times reasoning based on the image, we will provide them with corresponding images to let them finally confirm whether the sample needs to utilize multi-step image and text modalities reasoning to obtain the final multi-step reasoning paths.

\subsection{MCoT Domain Augmentation}
\label{sec:data-generation}

In order to make up for the missing data in previous work on mathematics and commonsense, we constructed \datasetname{} based on MATH~\cite{hendrycksmath2021} and Sherlock~\cite{hessel2022abduction} dataset to enhance the dataset within respective domains. More details are illustrated in Appendix~\ref{sec:appendix-generation}.

\noindent\textit{\textbf{Mathematics Domain Augmentation: }}
It is worth noting that MATH~\cite{hendrycksmath2021} is a single modal dataset only with textual questions, rationales and answers, lacking corresponding options and images.
To construct the options, we first prompt LLM to generate the related and similar options.
Then, for lack of images, we further convert the geometry code and formula code into images, and use HTML framework to splice them together.

\noindent\textit{\textbf{Commonsense Domain Augmentation: }}
In order to expand the field of commonsense, we use the Sherlock~\cite{hessel2022abduction}, which only contains some visual clues and does not contain any specific questions, options, and answers. 
Therefore, following \citet{zhang-etal-2023-crt}, we require LLM to cautiously generate questions, options, and answers. Specifically, we incorporate the multiple visual clues in Sherlock to LLM and enforce LLM generates models based on multiple image clues instead of single ones to ensure multi-step multi-modal reasoning.

\subsection{Quality Assurance}
\label{sec:quality}
This section aims to improve annotated data quality. More details are shown in Appendix~\ref{sec:appendix-quality}.

\noindent\textit{\textbf{Onboarding Test: }} Annotators are all required to undergo a preliminary test, annotating 100 samples. Their results are evaluated by three experts, and only those achieving at least 80\% accuracy proceed to subsequent annotation tasks.

\noindent\textit{\textbf{Human Annotation: }}
To address potential hallucinations or logical errors in generated samples, annotators are first asked to review and refine the rationale, ensuring accuracy and coherence.

\noindent\textit{\textbf{Human Recheck: }}
After that, these annotators are required to recheck all data twice to determine if the data meets multi-step multi-modal reasoning criteria and possesses a coherent logical rationale. All samples in \datasetname{} for which at least two annotators agree can be accepted. The kappa coefficient between annotators achieves 0.85, which indicates perfect agreement~\cite{landis1977measurement}.

\begin{figure}[t]
	\centering
	\includegraphics[width=0.47\textwidth]{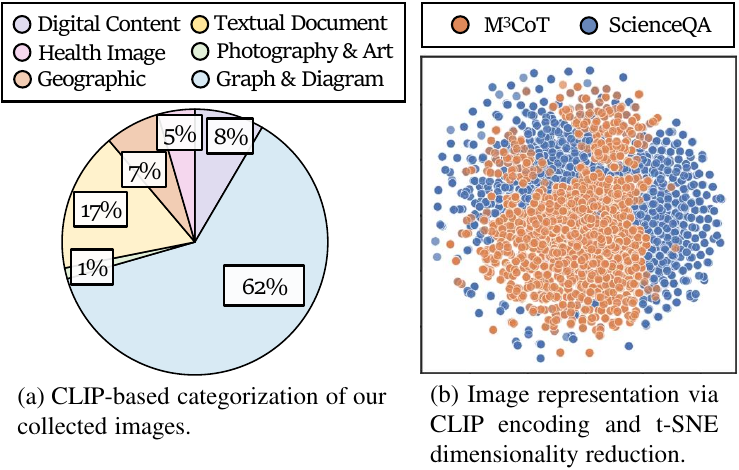}
	\caption{Image diversity analysis (a) and the representation comparison (b) between \datasetname{} and ScienceQA, where the point area in Figure (b) represents the image semantics coverage in the semantic space.
	}
	\label{fig:multimodal-diverse}
\end{figure}
\begin{figure}[t]
	\centering
	\includegraphics[width=0.46\textwidth]{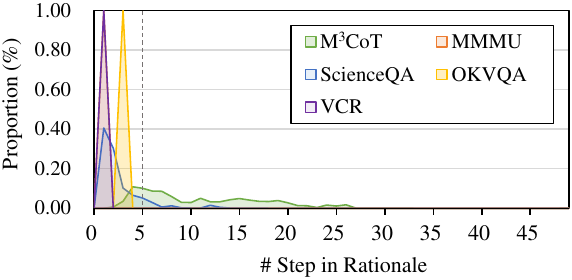}
	\caption{
		Comparison of the distribution of steps in the rationale for existing benchmarks. Notably, the distributions for MMMU and VCR overlap.
					}
	\label{fig:rationale_diverse}
\end{figure}
\begin{figure*}[t]
	\centering
	\includegraphics[width=0.97\textwidth]{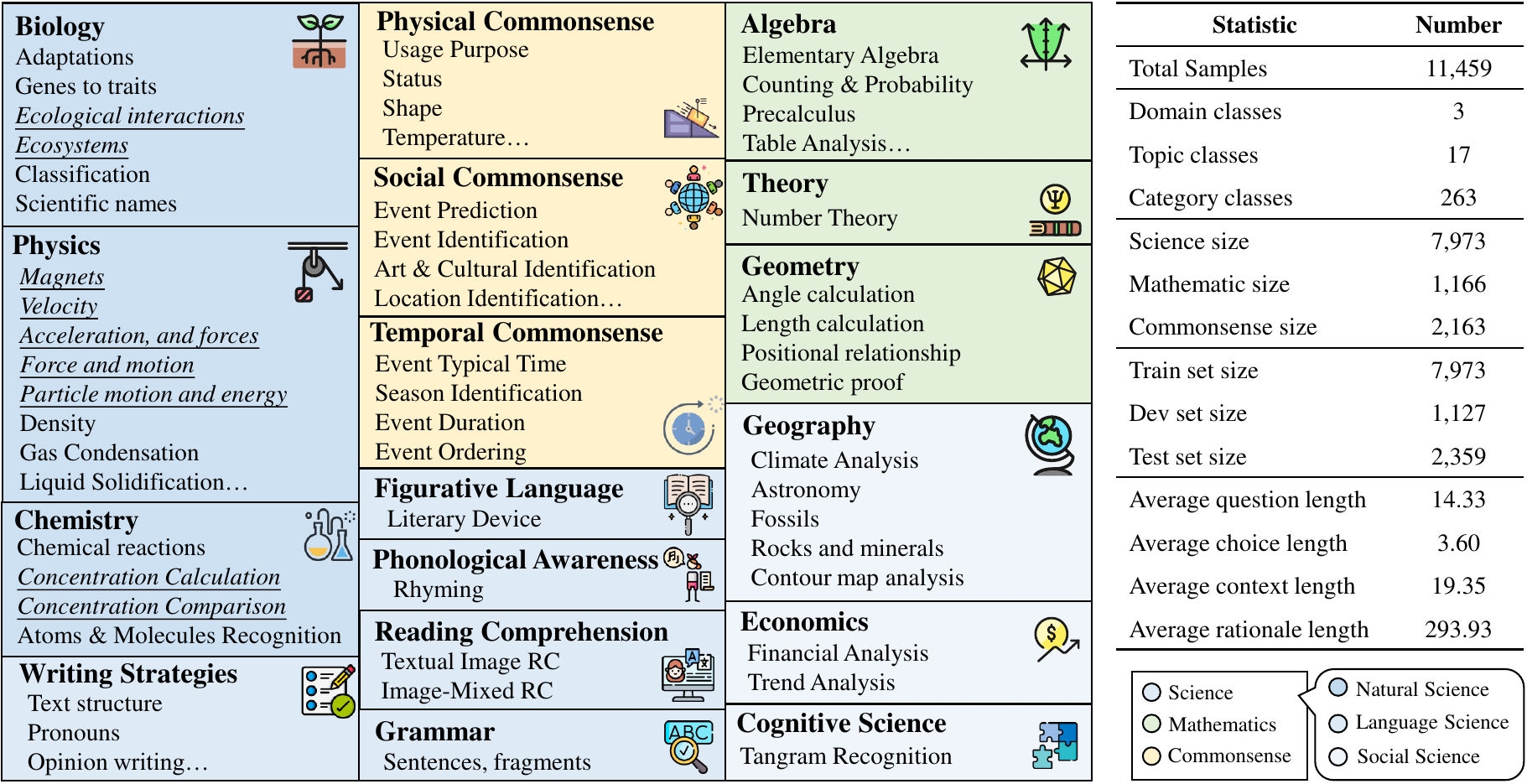}
	\caption{Detailed Analysis of \textbf{topic} and categories (partial) in the data set, where the \underline{\textit{underlined and italics}} mean the data we selected from ScienceQA.
	}
	\label{fig:datasets}
\end{figure*}
\section{Data Analysis}
This section provides some detailed data analysis to better understand \datasetname{}.

\noindent \textbf{Basic statistics }
\datasetname{} is partitioned randomly into three subsets: train, validation, and test splits, containing 7,863, 1,108, and 2,358 samples, respectively.
Compared to ScienceQA, \datasetname{} demands more intricate reasoning, with an average length of 294, much higher than ScienceQA's 48.

\noindent \textbf{Multi-modal diversity }
As shown in Figure~\ref{fig:multimodal-diverse} (a), \datasetname{} features diverse image types, which categorized by CLIP~\cite{radford2021learning}. Furthermore, Figure~\ref{fig:multimodal-diverse} (b) demonstrates that
\datasetname{} spans a broader semantic space, suggesting the enhanced semantic richness and coverage.

\noindent \textbf{Rationale diversity }
In comparison to existing benchmarks, the rationale process in \datasetname{} is characterized by an increased proportion of steps that are more uniformly distributed, as shown in Figure~\ref{fig:rationale_diverse}.
Specifically, ScienceQA averages 2.5 steps, OKVQA 3.0, MMMU only 1.0, and VCR only 1.0. In \datasetname{}, the reasoning process involves a significantly higher average of 10.9 steps,
highlighting the complexity and challenges presented by \datasetname{}.

\noindent \textbf{Domain Diversity }
In \datasetname{}, questions are categorized into three primary domains: science knowledge, mathematics and commonsense. As illustrated in Figure~\ref{fig:datasets}, the dataset encompasses 17 topics and 263 categories, highlighting the extensive variety of the questions. This variety is essential for assessing the generalization abilities of various models and for furthering multi-modal research.

\begin{table*}[t]
	\centering
	\begin{adjustbox}{width=0.97\textwidth}
		\begin{tabular}{lcccccccccc}
			\toprule
			\multirow{2}{*}{Model}  & \multicolumn{3}{c}{Science} & \multicolumn{3}{c}{Commonsense} & \multicolumn{3}{c}{Mathematics} & \multirow{2}{*}{Total}
			\\\cmidrule{2-10}
			& Lang &
			Natural & Social & Physical & Social & Temporal & Algebra & Geometry & Theory & 
			\\
			\midrule
			Random & 32.70 & 30.62 & 26.71 & 32.97 & 22.22 & 20.33 & 35.71 & 27.50 & 23.81 & 28.56 \\
																														\midrule
			\rowcolor{gray!8}\multicolumn{11}{c}{\textit{InstructBLIP-13B}~\cite{dai2023instructblip}}\\
			\midrule
			\texttt{Direct}~\cite{dai2023instructblip}  & 38.39 & 30.52 & 26.27 & 76.67 & 70.66 & 35.77 & 30.00 & 22.50 & 19.05 & 35.94 \\
			\texttt{CoT}~\cite{kojima2022large} & 38.39 & 30.01 & 27.55 & 80.00 & 70.25 & 33.33 & 30.71 & 21.25 & 19.05 & 36.07  \\
			\texttt{Desp-CoT}~\cite{wu2023role} & 16.59 & 27.84 & 22.77 & 54.44 & 52.89 & 30.08 & 27.86 & 28.75 & 28.57 & 29.25
			\\
			\texttt{CCoT}~\cite{mitra2023compositional} & 13.27 & 26.95 & 24.84 & 62.22 & 67.36 & 41.46 & 25.00 & 25.00 & 23.81 & 31.28
			\\
			\midrule
			\rowcolor{gray!8}\multicolumn{11}{c}{\textit{LLava-V1.5-13B}~\cite{liu2023improved}}\\
			\midrule
			\texttt{Direct}~\cite{liu2023improved}  & 36.97 & 27.46 & 20.22 & 52.22 & 23.55 & 27.64 & 22.86 & 45.00 & 4.76 & 27.05
			\\
			\texttt{CoT}~\cite{kojima2022large}& 46.45 & 38.31 & 27.87 & 67.78 & 64.05 & 49.59 & 26.43 & 30.00 & 23.81 & 39.52
			\\
			\texttt{Desp-CoT}~\cite{wu2023role} & 47.87 & 29.25 & 27.23 & 68.89 & 59.92 & 47.15 & 26.43 & 36.25 & 9.52 & 35.98
			\\
			\texttt{CCoT}~\cite{mitra2023compositional} & 38.86 & 31.55 & 28.18 & 72.22 & 61.57 & 39.84 & 29.29 & 36.25 & 28.57 & 36.45
			\\
			\midrule
			\rowcolor{gray!8}\multicolumn{11}{c}{\textit{CogVLM-17B}~\cite{wang2023cogvlm}}\\
			\midrule
			\texttt{Direct}~\cite{wang2023cogvlm} & 52.61 & 37.42 & 26.91 & 55.56 & 54.13 & 29.27 & 29.29 & 32.50 & 23.81 & 37.19
			\\
			\texttt{CoT}~\cite{kojima2022large} & 51.18 & 43.81 & 29.30 & 54.44 & 39.26 & 31.71 & 35.71 & 33.75 & 33.33 & 38.91
			\\
			\texttt{Desp-CoT}~\cite{wu2023role} & 46.92 & 35.63 & 25.80 & 48.89 & 47.52 & 38.21 & 27.14 & 31.25 & 19.05 & 35.07
			\\
			\texttt{CCoT}~\cite{mitra2023compositional} & 47.39 & 34.99 & 25.80 & 62.22 & 46.28 & 35.77 & 30.71 & 37.50 & 23.81 & 35.63
			\\
			\midrule
			\rowcolor{gray!8}\multicolumn{11}{c}{\textit{Gemini}~\cite{google2023gemini}}\\
			\midrule
			\texttt{Direct}~\cite{google2023gemini} & 73.93 & 41.25 & 31.21 & 56.67 & 71.49 & 62.60 & 30.71 & 27.50 & 28.57 & 45.17 \\
			\texttt{CoT}~\cite{kojima2022large} & 67.30 & 49.68 & 36.31 & 68.89 & 60.33 & 66.67 & 23.57 & 21.25 & 9.52 & 47.50 \\
			\texttt{Desp-CoT}~\cite{wu2023role} & 49.29 & 43.68 & 27.07 & 63.33 & 57.85 & 70.73 & 28.57 & 30.00 & 28.57 & 41.85 \\
			\texttt{CCoT}~\cite{mitra2023compositional} & 36.49 & 31.16 & 27.39 & 71.11 & 36.78 & 55.28 & 20.71 & 16.25 & 0.00 & 32.61 \\
			\midrule
			\rowcolor{gray!8}\multicolumn{11}{c}{\textit{GPT4V}~\cite{openai2023gpt4}}\\
			\midrule
			\texttt{Direct}~\cite{openai2023gpt4} & 80.09 & 54.66 & 43.95 & 87.78 & 67.77 & 82.11 & 42.14 & 43.75 & \textbf{42.86} & 56.95
			\\
			\texttt{CoT}~\cite{kojima2022large} & \textbf{90.52} & \textbf{63.09} & \textbf{46.97} & 83.33 & \textbf{75.21} & \textbf{82.93} & \textbf{45.71} & \textbf{50.00} & 38.10 & \textbf{62.60}
			\\
			\texttt{Desp-CoT}~\cite{wu2023role} & 79.62 & 54.66 & 36.94 & \textbf{88.89} & 74.38 & 73.98 & 20.71 & 32.50 & 33.33 & 53.54
			\\
			\texttt{CCoT}~\cite{mitra2023compositional} & 84.83 & 55.30 & 39.81 & 80.00 & 65.70 & 81.30 & 32.86 & 21.25 & 28.57 & 54.44
			\\
			\midrule
			Human & 97.63 & 91.70 & 87.92 & 97.80 & 94.24 & 91.87 & 85.71 & 90.00 & 76.19 & 91.17 \\
			\bottomrule
		\end{tabular}
	\end{adjustbox}
	\caption{
		Main experimental results on selected VLLMs. ``Random'' and ``Human'' performance are the average accuracy by three attempts. Detailed descriptions of these performances are shown in Appendix~\ref{append:heuristic-baselines}. Complete experiments are provided in Table~\ref{exp:all-exp}.
	}
	\label{exp:main-exp}
\end{table*}

%% file: sections/experiment.tex
\section{Experiments}

\begin{figure*}[t]
	\centering
	\includegraphics[width=0.99\textwidth]{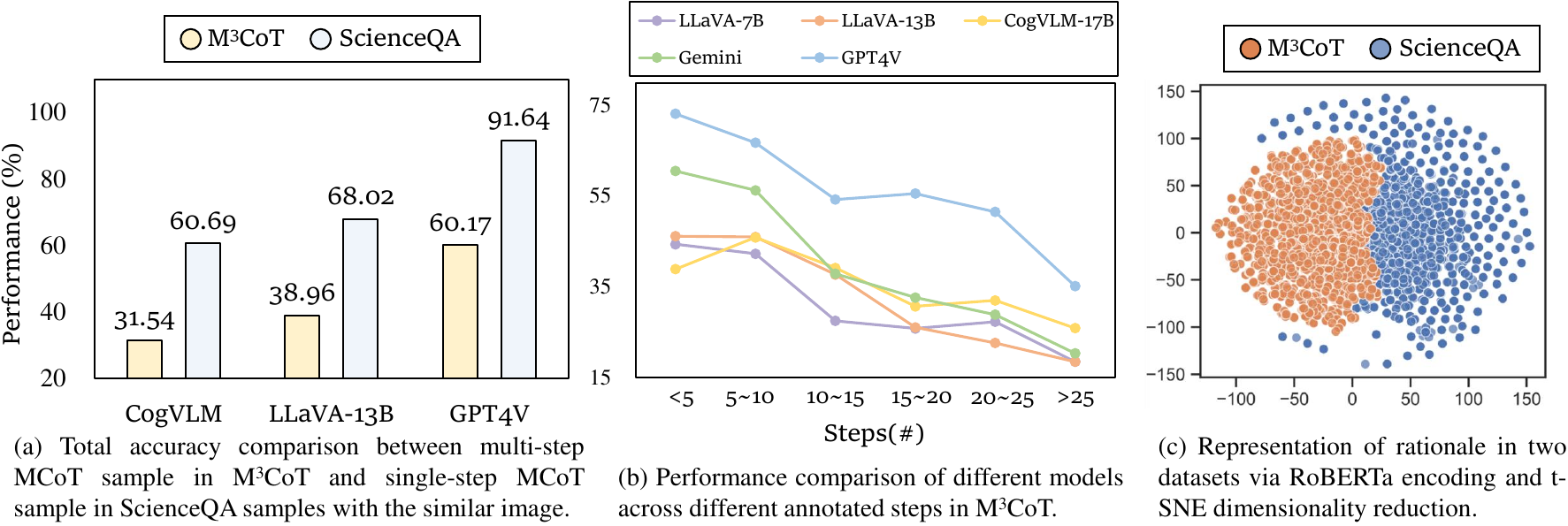}
	\caption{
				Performance comparison and reason analysis of \datasetname{} and ScienceQA.}
	\label{fig:comparison}
\end{figure*}
\subsection{Experiments Setting}
We evaluate various VLLMs in \datasetname{}, including \textit{Kosmos-2}~\cite{kosmos-2}, \textit{InstructBLIP}~\cite{dai2023instructblip}, \textit{LLaVA-V1.5}~\cite{liu2023improved}, \textit{CogVLM}~\cite{wang2023cogvlm}, \textit{Gemini}~\cite{google2023gemini}, \textit{GPT4V}~\cite{openai2023gpt4}. In addition, we explore some prompting strategies. Specifically, we utilize \texttt{Direct} approach to submitting samples in the VLLMs required format; \texttt{CoT}~\cite{kojima2022large} with ``Let's think step-by-step!''; \texttt{Desp-CoT}~\cite{wu2023role} with an initial image description prompting; \texttt{CCoT}~\cite{mitra2023compositional} with better description in graph format. Following the settings of \citet{kojima2022large,qin2023cross}, we extract the final generated answer through regular expressions.

\subsection{Results for \datasetname{}}
Results are presented in Table~\ref{exp:main-exp}. We have the following observations:

\paragraph{\textit{There remains a significant disparity between open source VLLMs and GPT4V}.}
Open source VLLMs still lag behind GPT4V by at least 7.98\% on the \datasetname{} benchmark. It highlights the limitations in the interaction and reasoning capabilities of existing open-source VLLMs, when compared to GPT-4V, especially in advanced tasks.

\paragraph{\textit{It exhibits a significant gap between GPT4V and human}.}
Despite GPT4V's impressive results, it substantially trails human performance,
demonstrating GPT4V still struggles to \datasetname{}.

\paragraph{\textit{Zero-shot Multi-modal Chain-of-Thought only benefits larger VLLMs}.}
As shown in Table~\ref{exp:main-exp} and Table~\ref{exp:all-exp}, MCoT strategy fails to enhance reasoning abilities in VLLM with fewer than 13B parameters.
Therefore, larger VLLMs ($\ge 13B$) can better observe emergent capabilities.

\subsection{Analysis}
\label{sec:analysis}
To gain a deeper understanding of why VLLMs fail on \datasetname{}, we analyze various factors to explore what influence the performance on \datasetname{}. We provide more analysis details
in  Appendix~\ref{sec:zero-cot-case} to confirm our speculations.
\begin{figure}[t]
	\centering
	\includegraphics[width=0.47\textwidth]{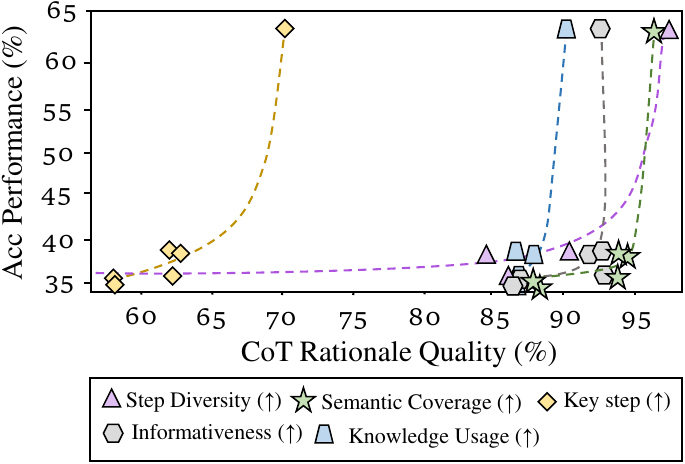}
	\caption{Analysis of the correlation between multi-dimensional qualities for model-generated rationale and final accuracy performance. The rationale qualities are computed by ROSCOE~\cite{golovneva2023roscoe}.}
	\label{fig:rationale-quality}
\end{figure}

\noindent\textbf{\textit{Multi-step MCoT poses a greater challenge than single-step one.}}
As shown in Figure~\ref{fig:comparison} (a), VLLM has achieved amazing performance in single-step reasoning. 
However, compared with single-step MCoT data in ScienceQA, multi-step MCoT data in \datasetname{} maintains at least a 29.06\% performance decrease (Figure~\ref{fig:comparison} (a)). In order to further understand the difference in model reasoning with different numbers of steps, we calculated the accuracy of different steps. As illustrated in Figure~\ref{fig:comparison} (b), an increase in the number of reasoning steps is associated with a significant decline in the model's performance.
In Figure~\ref{fig:comparison} (c), minimal rationale semantic distribution overlap between datasets further proves that the multi-step MCoT is an Out-of-Distribution (OOD) problem compared with single-step MCoT. For all, we attribute the low performance to the multi-step complexities for \datasetname{}.

\noindent\textbf{\textit{Multi-step MCoT needs higher rationale quality for better performance}.}
We comprehensively assess the predicted rationale quality of various VLLMs based on five dimension criteria.
As shown in Figure~\ref{fig:rationale-quality}, we observe that rationale quality incrementally improves \datasetname{} performance, while it markedly impacts the accuracy in CoT tasks. 
In the future, we believe that improving rational quality is one of the key challenges to solving \datasetname{}.

\begin{figure}[t]
	\centering
	\includegraphics[width=0.46\textwidth]{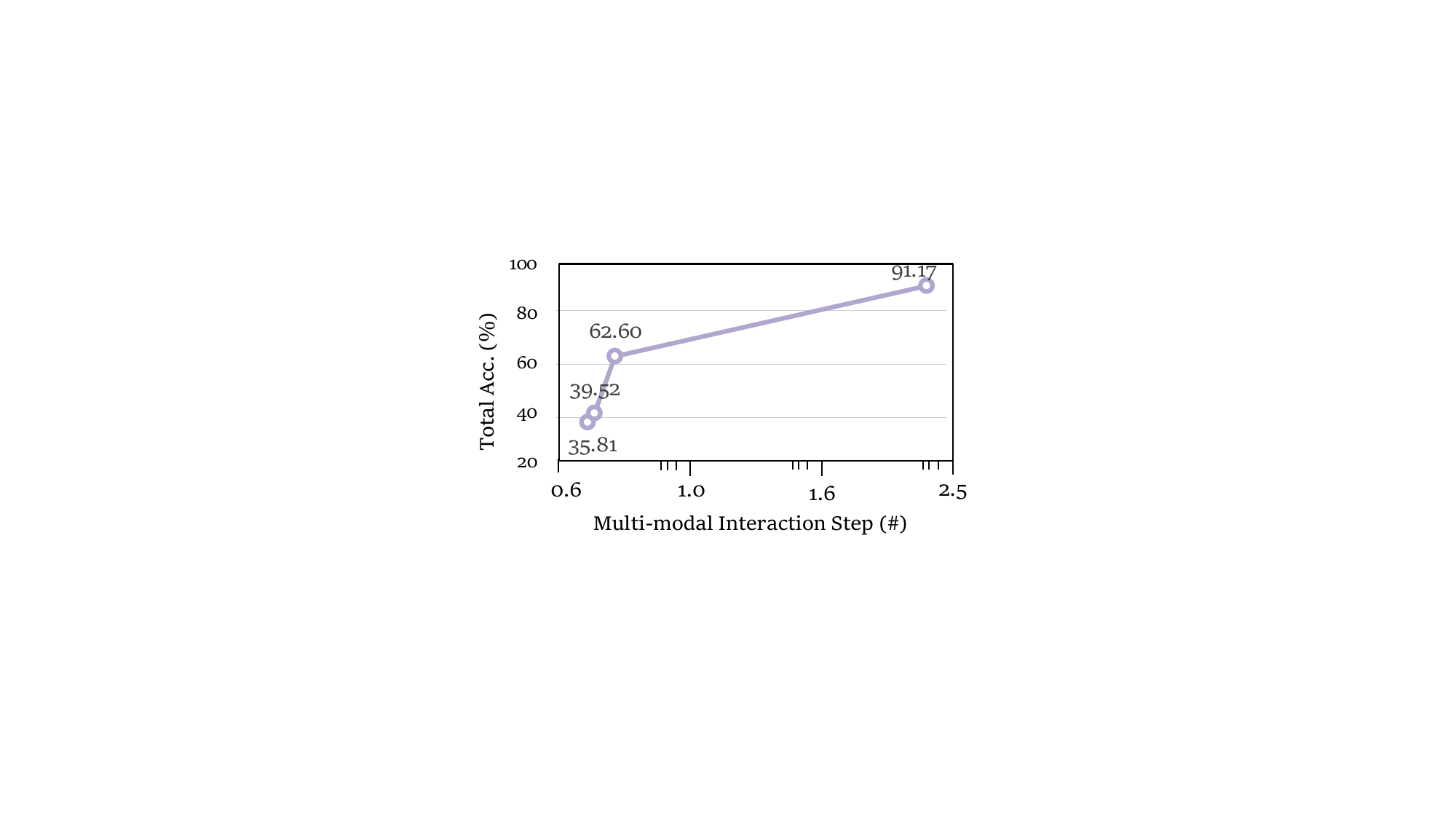}
	\caption{Analysis of the correlation between averaged multi-modal interaction steps and accuracy performance.}
	\label{fig:image-complex}
\end{figure}
\noindent\textbf{\textit{Multi-step MCoT needs more multi-modal interaction}.}
To assess the necessity for more complex multi-modal interaction reasoning in \datasetname{}, we examine how multi-modal interaction degrees impact performance.  Specifically,
we measure this by defining the similarity between images and reasoning steps to judge which steps are related to the image, identifying steps to sufficient similarity as multi-modal interaction steps. Figure~\ref{fig:image-complex} illustrates a positive correlation between averaged multi-modal interaction steps and reasoning performance, indicating \datasetname{} benefits from more multi-modal reasoning steps for optimal performance.
\subsection{Exploration}
In addition to the zero-shot CoT evaluation, we further evaluate the models on \datasetname{} under three setups: (1) \textit{Multi-modal Tool Usage}; (2) \textit{Multi-modal In-Context-Learning}; (3) \textit{Fine-tuning}. To our knowledge, these are the first comprehensive exploration of the multi-modal CoT scenarios.

\subsubsection{Tool Usage Exploration}
\paragraph{\textit{Multi-modal tool usage on text-modal LLM fails on \datasetname{}.}}
Several studies highlight that ChatGPT can well use external multi-modal tools to help multi-modal reasoning. However, Table~\ref{exp:lavr-exp} reveals that those tool-usage models in single modality are significantly worse than that of GPT4V by 28.21\%, and some even are worse than random baseline. 
We attribute it to the fact that the current tool usage framework cannot observe the visual modal during planning, which caused incorrect tool planning and tool usage, like confusing description and captioning tools.
(as shown in Appendix~\ref{sec:tool-case})
This indicates the necessity for enhanced multi-modal information interaction within \datasetname{}. 
And we will show more implementation  details in Appendix~\ref{append:tool}.

\begin{figure}[t]
	\centering
	\includegraphics[width=0.48\textwidth]{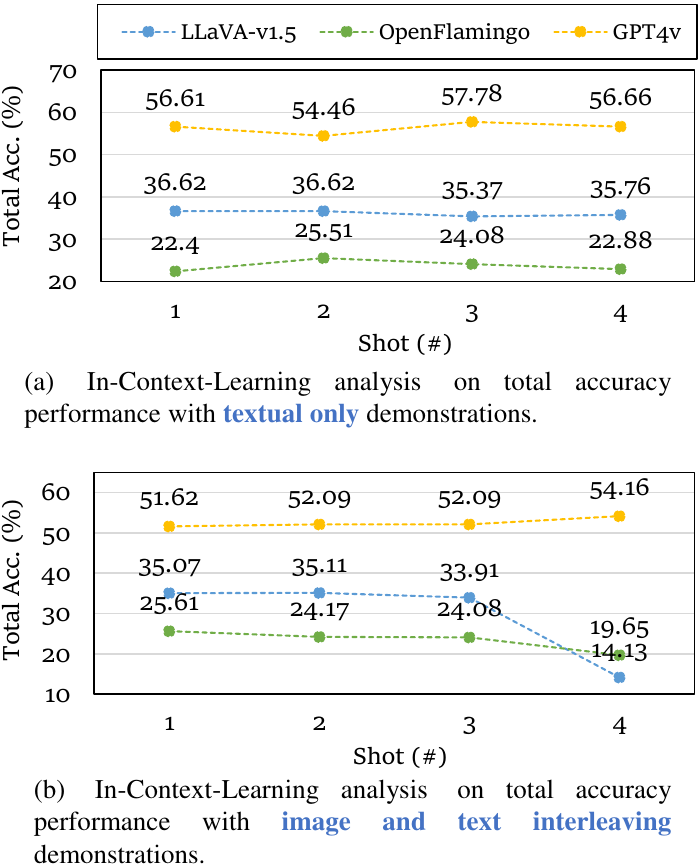}
	\caption{Performance change analysis of In-Context-Learning (ICL) CoT on textual modality and multi-modality demonstrations.}
	\label{fig:few-shot}
\end{figure}
\subsubsection{In-Context-Learning Exploration}

\noindent\textbf{\textit{Performance can not be boosted by text-only examples.}}
Contrasting with textual CoT~\cite{wei2022chain,shi2022language}, we find that ICL, even with chosen in-domain examples, fails to significantly improve multi-modal reasoning, as shown in Figure~\ref{fig:few-shot} (a). This suggests a need for more diverse multi-modal examples for \datasetname{}.

\noindent\textbf{\textit{Performance may even be harmed by image and text interleaving example.}}
In Figure~\ref{fig:few-shot} (b), it reveals that LLaVA-13B,  untrained in interleaved image-text data, suffers performance degradation with more samples. Surprisingly, despite being trained on interleaved image-text data, OpenFlamingo~\cite{awadalla2023openflamingo} still exhibits a slight performance decline.
In contrast, GPT4V, which is thoroughly trained on high-quality image-text interleaving examples, improves the performance as the number of shots increases. However, its performance is still lower than direct
\texttt{CoT}.
These indicate the future direction of high-quality interleaved samples and multi-step cross-modal interaction to enhance performance. All implementation details are shown in Appendix~\ref{append:icl}.

\begin{table*}[t]
	\centering
	\begin{adjustbox}{width=0.99\textwidth}
		\begin{tabular}{lcccccccccc}
			\toprule
			\multirow{2}{*}{Model}  & \multicolumn{3}{c}{Science} & \multicolumn{3}{c}{Commonsense} & \multicolumn{3}{c}{Mathematics} & \multirow{2}{*}{Total}
			\\\cmidrule{2-10}
			& Lang &
			Natural & Social & Physical & Social & Temporal & Algebra & Geometry & Theory & 
			\\
			\midrule
			Random & 32.70 & 30.62 & 26.71 & 32.97 & 22.22 & 20.33 & 35.71 & 27.50 & 23.81 & 28.56 \\
			\midrule
			\rowcolor{gray!8}\multicolumn{11}{c}{\textit{Tool-Usage}}\\
			\midrule
			{HuggingGPT}~\cite{shen2023hugginggpt} & 17.57 & 20.93 & 10.33 & 8.70 & 14.75 & 9.76 & 11.35 & 22.50 & 9.52 & 14.60 \\
			{VisualChatGPT}~\cite{wu2023visul} & 30.09 & 36.28 & 7.78 & 43.48 & 29.92 & 33.33 & 21.99 & 21.25 & 28.57 & 25.92 \\
			{IdealGPT}~\cite{you2023idealgpt} & 31.73 & 31.63 & 26.23 & 56.52 & 50.00 & 26.83 & 20.57 & 30.00 & 38.10 & 32.19 \\
			{Chameleon}~\cite{lu-2023-chameleon} & 43.87 & 26.05 & 25.44 & 39.13 & 37.30 & 48.78 & 17.73 & 26.25 & 23.81 & 34.29 \\
						\midrule
			\rowcolor{gray!8}\multicolumn{11}{c}{\textit{Finetuning (Traditional VLM)}}\\
			\midrule
			{MM-CoT}$_{base}$~\cite{zhang-2023-multi} & 41.71 & 46.49 & 39.90 & 59.34 & 60.91 & 27.64 & 48.57 & 35.00 & 28.57 & 44.85 \\
			{MC-CoT}$_{base}$~\cite{tan2023boosting} & 53.55 & 63.98 & 43.56 & 61.54 & 69.55 & 29.27 & 42.86 & 33.75 & 28.57 & 53.51 \\
			\hdashline
			{MM-CoT}$_{large}$~\cite{zhang-2023-multi} & 45.50 & 50.19 & 43.56 & 63.74 & 64.61 & 33.33 & 40.71 & 61.25 & 28.57 & 48.73 \\
			{MMR}~\cite{wei2023enhancing} & 50.24 & 50.32 & 43.56 & \textbf{76.92} & \textbf{66.67} & 31.71 & 50.71 & \textbf{65.00} & \textbf{38.10} & 50.67 \\
			{MC-CoT}$_{large}$~\cite{tan2023boosting} & \textbf{42.65} & \textbf{67.43} & \textbf{50.56} & 58.24 & 60.49 & \textbf{56.10} & \textbf{57.86} & 62.50 & 14.29 & \textbf{57.69} \\
			\midrule
			\rowcolor{gray!8}\multicolumn{11}{c}{\textit{Finetuning (VLLM)}}\\
			\midrule
			{LLaMA-Adaper-7B}~\cite{zhang2023llamaadapter} & 62.56 & 72.29 & 30.21 & 76.92 & 59.67 & 72.36 & 30.71 & 38.75 & 38.10 & 54.89 \\
			{LLaVA-V1.5-7B}~\cite{liu2023improved} & 65.88 & 73.44 & 35.14 & 80.22 & 56.79 & 67.48 & 32.86 & 47.50 & 19.05 & 56.74 \\
			{LLaVA-V1.5-13B}~\cite{liu2023improved} & \textbf{68.72} & 72.41 & \textbf{40.86} & \textbf{83.52} & 64.61 & 69.11 & \textbf{35.71} & 45.00 & 38.10 & \textbf{59.50} \\
			{CogVLM-17B}~\cite{wang2023cogvlm} & 65.88 & \textbf{77.52} & 29.09 & 81.32 & \textbf{65.43} & \textbf{75.61} & \textbf{35.71} & \textbf{46.25} & \textbf{47.62} & 58.25 \\
			\midrule
			{GPT4V}{$_{CoT}$}~\cite{openai2023gpt4}  & 90.52 & 63.09 & 46.97 & 83.33 & 75.21 & 82.93 & 45.71 & 50.00 & 38.10 & 62.60
			\\
			\midrule
			Human & 97.83 & 92.62 & 94.31 & 96.28 & 92.41 & 88.71 & 87.23 & 88.75 & 85.71 & 91.61 \\
			\bottomrule
		\end{tabular}
	\end{adjustbox}
	\caption{
		Fine-tuning results on various VLLMs.
	}
	\label{exp:lavr-exp}
\end{table*}

\subsubsection{Finetuning Exploration}
To further explore the improvement on \datasetname{}, we conduct finetuning experiments for more effective multi-modal reasoning. We will show more implementation details in Appendix~\ref{append:finetune}. 

\noindent\textbf{\textit{Finetuning on \datasetname{} can result better performance.}}
Table~\ref{exp:lavr-exp} reveals that our benchmark training set significantly enhances model performance. It enables traditional vision-language models (VLMs) to surpass the zero-shot VLLMs, which is the value of our dataset in boosting VLM effectiveness. Fine-tuned VLMs (the lowest is 44.85\%) outperform most open-source VLLMs with zero-shot prompting (the highest is 38.86\%). In addition, some fine-tuned VLMs have even surpassed Gemini's overall accuracy of 47.50\%, which demonstrates that fine-tuning can effectively boost the performance.

\noindent\textbf{\textit{Finetuning on VLLMs tends to be more effective than on Traditional VLM.}} Further, we found that the performance of VLLMs generally improves as their number of parameters increases. This also proves the importance of utilizing models with sufficient parameters in our \datasetname{} to achieve the target performance.

\begin{takeaways}
	\ \textbf{\textit{Takeaways:}} \textit{(1) Visual and textual information should be both considered for tool planning. (2) We should consider better multi-modal interleaving for better ICL in} \datasetname{}.
	\ \textit{(3) Fine-tuning has better hope on multi-step MCoT, compared with the failures of vanilla in-context-learning, tool usage, and prompting strategies.}
\end{takeaways}

%% file: sections/related.tex
\section{Related Work}

Chain-of-Thought~\cite{wei2022chain} (CoT) is a highly effective step-by-step strategy for enhancing zero-shot and few-shot reasoning in Large Language Models (LLMs)~\cite{kojima2022large,zhou2022least,zelikman2022star,qin2023cross,qin2024large,zhuang2023through}. 
In addition, some works begin to extend the textual CoT capabilities to multi-modal CoT reasoning (MCoT)~\cite{wangtowards,singh2023assessing,he2023multi}. 
To this end, \citet{lu2022learn} introduce the ScienceQA benchmark, laying the foundation for MCoT.
Subsequently, \citet{zhang2023multimodal} formally formalize the MCoT concept and enhanced its performance using a two-stage approach during multi-modal reasoning.
Additionally, \citet{wang2023t} develop a novel framework to integrate more knowledge with high-quality CoT rationales from larger LLMs for better MCoT reasoning.
Further, \citet{mondal2024kam} integrate CoT, Knowledge Graphs, and multi-modalities together for better MCoT.
\citet{ge2023chain,zheng2023ddcot,yao2023thinking} manually decouple the  chain-of-thought reasoning steps, integrating better multi-modal interaction.
Building upon these works, \citet{tan2023boosting} introduce the self-consistency mechanism~\cite{wang2022self} into the training process to enable more accurate/reliable reasoning.
\citet{wei2023enhancing} propose a novel approach to improve reasoning capabilities in image and text encoders through the integration of multi-hop cross-modal attention and sentence-level contrastive learning.
\citet{chen2023measuring} further extend the MCoT benchmark to generation tasks for better commonsense reasoning evaluation.

Compared with previous works, we first propose \datasetname{} to explore the multi-step MCoT and extend their application across a broader range of domains. 
In addition, we conduct comprehensive experiments on \datasetname{} and highlight some takeaways, facilitating future research.

%% file: sections/conclusion.tex
\section{Conclusion}

In this work, we introduce a novel benchmark (\datasetname{}), toward multi-domain, multi-step, and multi-modal chain-of-thought scenarios,
which is developed through a detailed and comprehensive process. 
In addition, we conduct a comprehensive analysis involving abundant multi-modal CoT methodologies to understand the limitation of existing frameworks on \datasetname{}. We sincerely aspire that our work can reassess existing advancements and inspire future research by highlighting new challenges and opportunities.

\section*{Limitations}
We introduce a new benchmark for multi-domain, multi-step, and multi-modal Chain of Thought (\datasetname{}) reasoning, performing in-depth analysis and exploring various CoT methodologies to better understand existing frameworks.
However, limited by unavoidable human subjectivity, manual annotation may introduce potential biases that may affect the reliability of the data.
Furthermore, with the advent of globalization, multilingualism has become increasingly important~\cite{qin2024multilingual}. However, due to regional and cost restrictions, the data does not take into account multilingual backgrounds and was developed only in the single language of English. Moreover, due to the possibility of discontinued or retired models, we will pay more attention to the open-source models in the future.

\section*{Ethical Considerations}
\paragraph{Data Access} We sourced our data from the ScienceQA~\cite{lu-2023-science}, MATH~\cite{hendrycksmath2021}, TabMWP~\cite{lu2022dynamic}, KiloGram~\cite{ji2022abstract}, and Sherlock~\cite{hessel2022abduction}. These are open-source and freely available for academic research, aligning with the commitment to ethical data use.
%\paragraph{Annotation Platform Construction} 

\paragraph{Participant Recruitment}
We recruit participants from universities and require all participants to pass the CET-6 exam or IELTS score of 6 or above. In addition, all participants come from all over China and may have some national biases. We blurred national differences in the data set as much as possible, limiting it to common human commonsense. All annotators gave informed consent and were compensated in excess of the local minimum wage. In addition, the site does not require IRB review.

\paragraph{Dataset Collection Process} 
Our annotation process begins with an onboarding test, introducing the task through 100 example questions. Participants are compensated \$20 for this initial phase, aimed at acquainting them with the task. Subsequently, annotators receive \$15 per hour for their contributions, accumulating approximately 450 human-hours for manual annotations.
Following this, a recheck process to ensure correct labeling is added an additional 60 hours of work.
Overall, six experts and three students are engaged to fulfill the annotation and recheck tasks.

\section*{Acknowledgments}
This work was supported by the National Natural Science Foundation of China (NSFC) via grant 62306342, 62236004 and 62206078. This work was also sponsored by CCF-Baidu Open Fund and Excellent Young Scientists Fund in Hunan Province (2024JJ4070). We are grateful for resources from the High Performance Computing Center of Central South University.

%% file: sections/appendix.tex
\begin{table*}
	\centering
	\begin{adjustbox}{width=0.97\textwidth}
		\begin{tabular}{lcccccccccc}
			\toprule
			\multirow{2}{*}{Model}  & \multicolumn{3}{c}{Science} & \multicolumn{3}{c}{Commonsense} & \multicolumn{3}{c}{Mathematics} & \multirow{2}{*}{Total}
			\\\cmidrule{2-10}
			& Lang &
			Natural & Social & Physical & Social & Temporal & Algebra & Geometry & Theory & 
			\\
			\midrule
			Random & 32.70 & 30.62 & 26.71 & 32.97 & 22.22 & 20.33 & 35.71 & 27.50 & 23.81 & 28.56 \\
			\midrule
			\rowcolor{gray!8}\multicolumn{11}{c}{\textit{Kosmos-2-2B}~\cite{kosmos-2}}\\
			\midrule
			\texttt{Direct}~\cite{kosmos-2} & 10.43 & 28.61 & 21.18 & 33.33 & 17.77 & 28.46 & 21.43 & 21.25 & 14.29 & 23.17
			\\
			\texttt{CoT}~\cite{kojima2022large} & 18.48 & 24.14 & 14.65 & 30.00 & 14.46 & 9.76 & 17.14 & 18.75 & 0.00 & 18.68
			\\
			\texttt{Desp-CoT}~\cite{wu2023role} & 0.00 & 0.00 & 0.00 & 1.11 & 0.00 & 0.00 & 0.00 & 0.00 & 0.00 & 0.04
			\\
			\texttt{CCoT}~\cite{mitra2023compositional} & 0.00 & 0.00 & 0.16 & 2.22 & 7.44 & 1.63 & 0.00 & 0.00 & 0.00 & 0.99
			\\
						\midrule
			\rowcolor{gray!8}\multicolumn{11}{c}{\textit{InstructBLIP-7B}~\cite{dai2023instructblip}}\\
			\midrule
			\texttt{Direct}~\cite{dai2023instructblip} & 30.81 & 32.31 & 27.55 & 60.00 & 66.94 & 39.02 & 35.71 & 31.25 & 33.33 & 36.11
			\\
			\texttt{CoT}~\cite{kojima2022large}& 38.39 & 30.01 & 26.43 & 80.00 & 70.25 & 33.33 & 30.71 & 21.25 & 19.05 & 35.76
			\\
			\midrule
			\rowcolor{gray!8}\multicolumn{11}{c}{\textit{InstructBLIP-13B}~\cite{dai2023instructblip}}\\
			\midrule
			\texttt{Direct}~\cite{dai2023instructblip}  & 38.39 & 30.52 & 26.27 & 76.67 & 70.66 & 35.77 & 30.00 & 22.50 & 19.05 & 35.94 \\
			\texttt{CoT}~\cite{kojima2022large} & 38.39 & 30.01 & 27.55 & 80.00 & 70.25 & 33.33 & 30.71 & 21.25 & 19.05 & 36.07  \\
			\texttt{Desp-CoT}~\cite{wu2023role} & 16.59 & 27.84 & 22.77 & 54.44 & 52.89 & 30.08 & 27.86 & 28.75 & 28.57 & 29.25
			\\
			\texttt{CCoT}~\cite{mitra2023compositional} & 13.27 & 26.95 & 24.84 & 62.22 & 67.36 & 41.46 & 25.00 & 25.00 & 23.81 & 31.28
			\\
			\midrule
			\rowcolor{gray!8}\multicolumn{11}{c}{\textit{LLava-V1.5-7B}~\cite{liu2023improved}}\\
			\midrule
			\texttt{Direct}~\cite{liu2023improved} & 43.13 & 37.16 & 26.43 & 66.67 & 58.26 & 30.89 & 22.14 & 35.00 & 14.29 & 36.63 \\
			\texttt{CoT}~\cite{kojima2022large}& 38.86 & 33.59 & 25.48 & 71.11 & 65.29 & 39.02 & 29.29 & 16.25 & 4.76 & 35.81 \\
			\texttt{Desp-CoT}~\cite{wu2023role} & 34.12 & 32.18 & 25.32 & 65.56 & 57.85 & 41.46 & 24.29 & 31.25 & 28.57 & 34.43
			\\
			\texttt{CCoT}~\cite{mitra2023compositional} & 26.54 & 35.50 & 28.66 & 62.22 & 55.79 & 44.72 & 29.29 & 31.25 & 9.52 & 35.72
			\\
			\midrule
			\rowcolor{gray!8}\multicolumn{11}{c}{\textit{LLava-V1.5-13B}~\cite{liu2023improved}}\\
			\midrule
			\texttt{Direct}~\cite{liu2023improved}  & 36.97 & 27.46 & 20.22 & 52.22 & 23.55 & 27.64 & 22.86 & 45.00 & 4.76 & 27.05
			\\
			\texttt{CoT}~\cite{kojima2022large}& 46.45 & 38.31 & 27.87 & 67.78 & 64.05 & 49.59 & 26.43 & 30.00 & 23.81 & 39.52
			\\
			\texttt{Desp-CoT}~\cite{wu2023role} & 47.87 & 29.25 & 27.23 & 68.89 & 59.92 & 47.15 & 26.43 & 36.25 & 9.52 & 35.98
			\\
			\texttt{CCoT}~\cite{mitra2023compositional} & 38.86 & 31.55 & 28.18 & 72.22 & 61.57 & 39.84 & 29.29 & 36.25 & 28.57 & 36.45
			\\
			\midrule
			\rowcolor{gray!8}\multicolumn{11}{c}{\textit{CogVLM-17B}~\cite{wang2023cogvlm}}\\
			\midrule
			\texttt{Direct}~\cite{wang2023cogvlm} & 52.61 & 37.42 & 26.91 & 55.56 & 54.13 & 29.27 & 29.29 & 32.50 & 23.81 & 37.19
			\\
			\texttt{CoT}~\cite{kojima2022large} & 51.18 & 43.81 & 29.30 & 54.44 & 39.26 & 31.71 & 35.71 & 33.75 & 33.33 & 38.91
			\\
			\texttt{Desp-CoT}~\cite{wu2023role} & 46.92 & 35.63 & 25.80 & 48.89 & 47.52 & 38.21 & 27.14 & 31.25 & 19.05 & 35.07
			\\
			\texttt{CCoT}~\cite{mitra2023compositional} & 47.39 & 34.99 & 25.80 & 62.22 & 46.28 & 35.77 & 30.71 & 37.50 & 23.81 & 35.63
			\\
			\midrule
			\rowcolor{gray!8}\multicolumn{11}{c}{\textit{Gemini}~\cite{google2023gemini}}\\
			\midrule
			\texttt{Direct}~\cite{google2023gemini} & 73.93 & 41.25 & 31.21 & 56.67 & 71.49 & 62.60 & 30.71 & 27.50 & 28.57 & 45.17 \\
			\texttt{CoT}~\cite{kojima2022large} & 67.30 & 49.68 & 36.31 & 68.89 & 60.33 & 66.67 & 23.57 & 21.25 & 9.52 & 47.50 \\
			\texttt{Desp-CoT}~\cite{wu2023role} & 49.29 & 43.68 & 27.07 & 63.33 & 57.85 & 70.73 & 28.57 & 30.00 & 28.57 & 41.85 \\
			\texttt{CCoT}~\cite{mitra2023compositional} & 36.49 & 31.16 & 27.39 & 71.11 & 36.78 & 55.28 & 20.71 & 16.25 & 0.00 & 32.61 \\
			\midrule
			\rowcolor{gray!8}\multicolumn{11}{c}{\textit{GPT4V}~\cite{openai2023gpt4}}\\
			\midrule
			\texttt{Direct}~\cite{openai2023gpt4} & 80.09 & 54.66 & 43.95 & 87.78 & 67.77 & 82.11 & 42.14 & 43.75 & \textbf{42.86} & 56.95
			\\
			\texttt{CoT}~\cite{kojima2022large} & \textbf{90.52} & \textbf{63.09} & \textbf{46.97} & 83.33 & \textbf{75.21} & \textbf{82.93} & \textbf{45.71} & \textbf{50.00} & 38.10 & \textbf{62.60}
			\\
			\texttt{Desp-CoT}~\cite{wu2023role} & 79.62 & 54.66 & 36.94 & \textbf{88.89} & 74.38 & 73.98 & 20.71 & 32.50 & 33.33 & 53.54
			\\
			\texttt{CCoT}~\cite{mitra2023compositional} & 84.83 & 55.30 & 39.81 & 80.00 & 65.70 & 81.30 & 32.86 & 21.25 & 28.57 & 54.44
			\\
			\midrule
			Human & 97.63 & 91.70 & 87.92 & 97.80 & 94.24 & 91.87 & 85.71 & 90.00 & 76.19 & 91.17 \\
			\bottomrule
		\end{tabular}
	\end{adjustbox}
	\caption{
		The overall experimental results using selected VLLMs by zero-shot prompting. The ``\texttt{Direct}'' approach refers to submitting samples in the VLLM required format. ``Human'' is the average accuracy achieved by three college students who have successfully completed a relevant assessment. Complete experiments are provided in the Appendix.
	}
	\label{exp:all-exp}
\end{table*}
\section*{Appendix}
\begin{figure*}[t]
	\centering
	\includegraphics[width=0.97\textwidth]{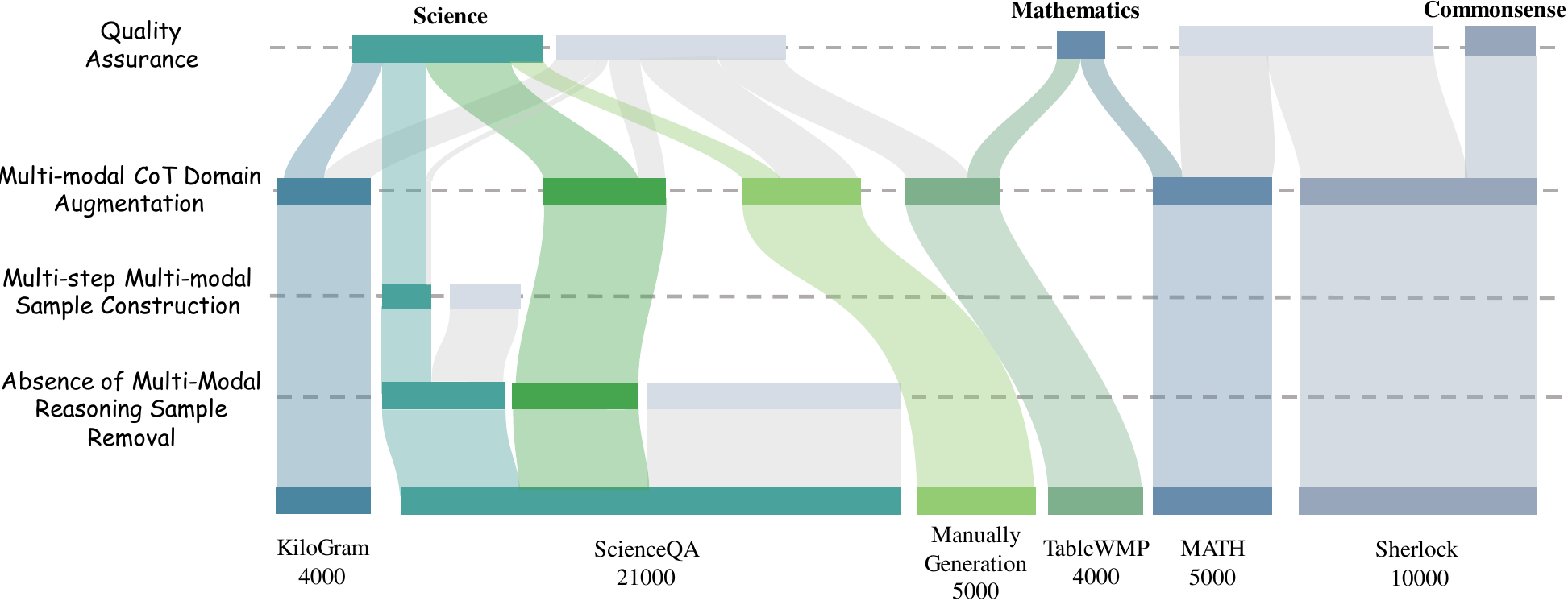}
	\caption{Sample distribution flow chart retained, generated, discarded at different stages
	}
	\label{fig:annotaion}
\end{figure*}
\section{Dataset Annotation Details}
\subsection{Statistical Analysis of Existing Datasets}
\label{append:exist-data}
In this study, we conduct a comprehensive analysis of the prevalence of multi-step reasoning within existing datasets. The analysis focuses on the "MMCoT" column in Figure~\ref{exp:data_comparison}(b), which represents the proportion of multi-step multi-modal CoT (Chain of Thought) data. This examination is critical as it reveals a significant deficiency in the current datasets regarding multi-step reasoning capabilities.
Our findings indicate that at least 79\% of the data across all the benchmarks we examined lack sufficient multi-step reasoning elements. This highlights a pervasive issue in the design and utilization of these datasets, where the absence of complex reasoning processes could impede the development of more sophisticated multi-modal models.

The term \%MMCoT, as used in this context, is consistent with the representation in Figure~\ref{exp:data_comparison}(b). Both denote the proportion of multi-step multi-modal CoT data. To ensure the accuracy of our analysis, we employed a rigorous sampling method. We selected a stratified random sample consisting of 20\% of the dataset, evenly distributed, for manual inspection. This approach allowed us to precisely determine the proportion of multi-step multi-modal CoT data within the existing datasets.

The manual inspection involved a detailed review of each sampled data point to identify and categorize the presence of multi-step reasoning. This meticulous process ensured that our findings were not only statistically significant but also reflective of the true nature of the datasets. The results underscore the necessity for enhanced dataset designs that incorporate more multi-step reasoning tasks, thereby facilitating the development of advanced multi-modal models capable of handling complex reasoning scenarios.

In conclusion, our statistical analysis sheds light on a critical gap in existing datasets, emphasizing the need for more comprehensive data that can better support the advancement of multi-modal reasoning capabilities. %This study provides a foundational understanding that can guide future dataset development and model training strategies in the field of multi-modal artificial intelligence.

\subsection{The Details of Absence of Visual Modal Reasoning Sample Removal}
\label{sec:annotate-first}
We develop the annotation interface based on the open-source Gradio framework. We segment the dataset, distribute the scripts, and deploy them to local computers.
In addition, we have designed some manual guidelines for annotators. 
These guidelines need to be followed by annotation experts. Specifically, our method flow is as follows:
\begin{itemize}
	\item[1. ] We will first mask the image so that the expert can only see the text modal questions, options, rationale, and answers.
	\item[2. ] The expert needs to directly judge whether the question can directly infer the rationale. If it cannot be inferred, it may be necessary for the visual modal information for the rationale generation.
	\item[3. ] And then we ask the experts to check the image to confirm this.
\end{itemize}

For each step, the guideline instructions are as follows:

\begin{mybox}
	\ \ 
	
	\textbf{[Instruction$_1$]}
	
	Firstly, we will conceal all visual content, allowing you to concentrate exclusively on the textual elements, including the questions, available options, correct answers, and rationales.
	
	You need to follow these instructions for annotation:
	\begin{itemize}
		\item You should determine whether the text alone can convey the full context and solution logic of the question.
		\item You should evaluate whether the textual information is sufficient for a comprehensive understanding and rationale formulation.
	\end{itemize}
	
	\textbf{[Instruction$_2$]}
	
	In cases where the text-based evaluation indicates a need for visual information, please click the ``Image Display'' button. 
	
	Then, the image will be reintroduced for further recheck, confirming the requirement for a multi-modal approach in certain scenarios.
	
	\ 
\end{mybox}

We used three experts to conduct majority voting to judge this matter. This part removes at least 30\% of ScienceQA's multi-modal sample, which also illustrates the limitations of the existing data.

\subsection{The Details of Multi-step MCoT Sample Construction}
\label{sec:annotate-second}
\noindent\textbf{\textit{Automatic Sample Removal: }}
In this step, we automatically filter out samples with overly simplistic rationales comprising fewer than two steps. This process reduces the manual annotation burden and enhances the reliability of \datasetname{}. Since the rationale in ScienceQA includes at least one conclusion and one step of reasoning, samples with fewer than two steps indicate that multiple visual cues were not used for MMCoT reasoning. Thus, this filtering step minimizes annotation workload and costs. Notably, samples with multi-step and single-step reasoning still require manual evaluation in our study.

\subsection{Domain Augmentation Details}
\subsubsection{Mathematics Domain Augmentation Details}
Firstly, the MATH~\cite{hendrycksmath2021} dataset consists solely of textual mathematical questions, complete with detailed rationales and answers. However, it does not include multiple-choice options and illustrative images, limiting its utility in \datasetname{}.

To address this, we employ \texttt{gpt-3.5-turbo} to generate relevant and similar multiple-choice options for each question, enhancing the dataset's versatility. 
Specifically, for the option generation phase, we prompt the LLM with specific questions from the dataset, followed by instructions to generate four plausible options, one correct and three distractors. The prompt included guidelines for ensuring the options are closely related to the question's content, challenging yet not misleading.
Specifically, the prompt is defined as follows:

\begin{mybox}
	\ \ 
	
	\textbf{[Input]}
	
	Here is the mathematical question:
	
	\textit{<Context, Question, Answer, Option>}\vspace{5pt}
		
\ 
\end{mybox}
\begin{mybox}
\ \ 

	\textbf{[Instruction]}
	
	Generate four multiple-choice options: one correct answer and three plausible but incorrect distractors. 

	Ensure the distractors are relevant and challenging without being misleading. 
	
	The options should closely relate to the question's subject matter and provide a meaningful test of the reader's understanding.
	
	\ 
\end{mybox}

To compensate for the absence of visual content, we developed a method to translate mathematical expressions and geometric figures described in the questions into visual representations. This process involved generating PNG images from the mathematical and geometric codes. Subsequently, we used a combination of HTML and CSS to integrate these images with the textual content, creating a cohesive multi-modal dataset.

\subsubsection{Commonsense Domain Augmentation Details}
\label{sec:appendix-generation}
In our study, we aim to enhance commonsense reasoning domain for \datasetname{} by leveraging visual clues from the Sherlock~\cite{hessel2022abduction} dataset. Unlike traditional vision question answering datasets, Sherlock~\cite{hessel2022abduction} does not include predefined questions, options, or answers, focusing instead on visual information to stimulate inference and deduction.

Recent advancements in LLMs have showcased their potential in generating diverse and contextually relevant data~\cite{zhang-etal-2023-crt}. Building on this, our approach involves prompting LLMs with visual clues from Sherlock, tasking them to generate coherent and contextually appropriate questions, multiple-choice options, and corresponding answers. This process demands careful design to ensure the prompts effectively communicate the visual information and desired output format to the LLM.

To ensure comprehensive multi-step, multi-modal reasoning, we develop a prompting methodology to trigger LLMs to consider multiple visual clues simultaneously.
Specifically, our experimental setup includes detailed prompting strategies that describe the visual clues from a structured manner to natural language description, allowing the LLM to understand and interpret the information accurately.
The prompt is defined as follows:

\begin{mybox}
	\ \ 
	
	\textbf{[Instruction]}
	
	Given the visual clues, generate a question that requires commonsense reasoning to answer.
	Then, provide four options (A, B, C, D), one of which is the correct answer, and the others are plausible distractors.\vspace{5pt}
	
	\textbf{[Highlighting]}

	Ensure that the question and options leverage insights from at least two visual clues for deeper multi-modal interactions.\vspace{5pt}

	\textbf{<One-shot Example>}\vspace{5pt}

	\textbf{[Visual Clue]}
	
	\textit{<multiple detailed description of the visual clues>}
	
	\ 
\end{mybox}
\noindent where \textbf{[One-shot Example]} denotes we used one-shot in-context-learning to allow the model to better learn the generation of related samples.

In addition, for some topics in ScienceQA where the data is too sparse due to the last ``absence of multi-modal reasoning sample removal'' and ``multi-step multi-modal sample construction'', we used a similar method to synthesize data in ScienceQA or used other open source data sets, like TabMWP~\cite{lu2022dynamic}, KiloGram~\cite{ji2022abstract}, for data augmentation. Moreover, we also manually synthesized some geographical images through Matplotlib, constructed some samples using rules, and polished and modified them using ChatGPT.

\subsection{Quality Assurance Details}
\label{sec:appendix-quality}
Due to space limitations, we only describe the rationale rewriting ($\S$\ref{sec:appendix-rewriting}) section in the appendix. This part is mainly to improve the quality of the rationale data set.
\subsubsection{Human Annotation Details}
In order to better mark the correctness of the logical chain of reasoning, first, we divide the steps according to the ROSCOE~\cite{golovneva2023roscoe} settings to obtain a clearer step-by-step rationale visualization. Secondly, 
we provide the corresponding sample image, question, options, answer and step-segmented rationale for experts to annotate each time. During annotation, we allow experts to discard samples that are of poor quality and cannot be modified to ensure the quality of the data set.

\subsubsection{Human Recheck Details}
In assessing the capability of a given sample to fulfill the criteria for multi-step multi-modal reasoning, this process employs a structured approach. Initially, we decompose the reasoning rationale into discrete steps by ROSCOE~\cite{golovneva2023roscoe}. Following this segmentation, experts are required to focus more on ascertaining which step needs image modality in rationale. After that, it requires further verification that the sample has considered integration of image modality for a minimum of two distinct steps. This methodological framework ensures a thorough recheck of the sample with the specified requirements of multi-step multi-modal reasoning. Specifically, the instructions given by our experts are as follows:
\begin{mybox}
	\ \ 
	
	\textbf{[Input]}
	
	Here is an example: \textit{<EXAMPLE>}\vspace{5pt}
	
	\textbf{[Instruction]}

	You need to judge whether a given sample meets the requirements of multi-step multi-modal reasoning.\vspace{-2pt}
	\begin{itemize}
		\setlength{\itemsep}{3pt}
		\setlength{\parsep}{0pt}
		\setlength{\parskip}{0pt}
		\item[1.] First, we have broken down the steps for you.
		\item[2.] Secondly, please determine which steps in rationale require image modality.
		\item[3.] Finally, please confirm whether the sample requires image modality for at least two steps.
	\end{itemize}
	\ 
\end{mybox}
\noindent where ``\textit{[EXAMPLE]}'' represents an example containing an image, question, options, answer, step-segmented rationale and annotation detail information.

Furthermore, the human recheck process actually has two rounds. In the second round, the sample discard rate is less than 5\%.

\subsubsection{Rationale Rewriting}
\label{sec:appendix-rewriting}
The rationale quality within the ScienceQA dataset has been found poor expression, with some explanations not adequately addressing the posed questions. To mitigate this issue, we have employed \texttt{gpt-3.5-turbo} to perform rationale rewriting, aiming to elevate the overall quality of \datasetname{} before human annotation.
Specifically, to achieve this, we designed a specific prompting strategy for the LLM, which prompt is defined as follows:

\begin{mybox}
	\ \ 
	
	\textbf{[Instruction]}
	
	Improve the quality of ScienceQA dataset rationales by rewriting them for enhanced relevance, accuracy, and clarity.\vspace{-4pt}
	\begin{itemize}
		\setlength{\itemsep}{3pt}
		\setlength{\parsep}{0pt}
		\setlength{\parskip}{0pt}
	\item[1. ] Read the question and the provided rationale from the ScienceQA dataset.
	\item[2. ] Evaluate the existing rationale for its relevance and accuracy in answering the question.
	\item[3. ] Rewrite the rationale to better answer the question, ensuring the new version is clear, concise, and directly related to the question's core topic. Maintain scientific accuracy and use accessible language suitable for the intended audience.
	\end{itemize}\vspace{-2pt}

	\textbf{[One-shot Example]}

	Question: \textit{<Question>}
	
	Original Rationale: \textit{<Rationale>}
		
	Rewritten Rationale: \textit{<Rewritten Rationale>}\vspace{5pt}

	\textbf{[Input]}
	
	Question: \textit{<Question>}
	
	Original Rationale: \textit{<Rationale>}
	
	Rewritten Rationale:
	
	\ 
\end{mybox}

This approach ensures that the rewritten rationales are not only relevant but also adhere to a high standard of clarity and coherence. Each rationale is assessed both automatically and manually to confirm its relevance and quality improvement over the original version.

\subsection{Image Redundancy Removal}
Additionally, we observed a significant number of highly similar samples in the ScienceQA.
To reduce redundancy and maintain diversity for image, we remove samples where the questions are identical, and the grayscale image similarity exceeded 99\%.

\section{Experiment Details}
\subsection{Main Result Details}
\label{append:main-result}
Due to space limitations, we only show some of the LLM test results in the main table. The specific experimental results are shown in Table~\ref{exp:all-exp}.
\subsubsection{Heuristic baselines}
\label{append:heuristic-baselines}
This study employs two heuristic baselines. The first is a random selection method, where an answer is chosen randomly, with its accuracy determined by averaging three random seeds. The second baseline evaluates human performance through participants who must successfully complete preliminary qualification tasks. The ``Human'' accuracy is the average accuracy achieved by three participants.

\subsection{Exploration Details}
\subsubsection{Tool Usage Details}
\label{append:tool}
\paragraph{Model Selection}
In this section, we introduce a suite of tool-augmented LLMs, including \textit{HuggingGPT}~\cite{shen2023hugginggpt}, \textit{VisualChatGPT}~\cite{wu2023visul}, \textit{IdealGPT}~\cite{you2023idealgpt}, and \textit{Chameleon}~\cite{lu-2023-chameleon}.
Specifically, \textit{VisualChatGPT}~\cite{wu2023visul} and \textit{IdealGPT}~\cite{you2023idealgpt} are engineered to tackle complex issues through iterative problem-solving processes. Conversely, \textit{HuggingGPT}~\cite{shen2023hugginggpt}, and \textit{Chameleon}~\cite{lu-2023-chameleon}
employ LLMs to decompose complicated challenges into a series of manageable sub-problems, addressing them in a sequential manner. This array of approaches highlights the diverse capabilities and potential of LLM-aided visual reasoning systems in executing sophisticated problem-solving strategies.

\subsubsection{In-Context-Learning Details}
\label{append:icl}
\paragraph{Model Selection}
In this section, we explore three notable models: \textit{LLaVA-V1.5-13B}~\cite{liu2023improved}, \textit{OpenFlamingo-7B}~\cite{awadalla2023openflamingo}, and \textit{GPT4V}~\cite{openai2023gpt4}, each demonstrating unique capabilities in the context of In-Context Learning (ICL). LLaVA-V1.5-13B~\cite{liu2023improved} is built upon a foundation of instruction-following data of high quality without any specific image-text interleaving training. \textit{OpenFlamingo}~\cite{awadalla2023openflamingo} is a VLLM, optimized for tasks involving complex image-text interleaving sequences.
GPT4V~\cite{openai2023gpt4} is the state-of-the-art VLLM that can learn efficiently from limited image-text interleaving demonstrations. Therefore, we assume it is well-trained in image-text interleaving scenarios.

\paragraph{Exemplar Selection}
In order to complete in-domain sample selection as much as possible, we only randomly select samples under the same categories from the development set.

\begin{figure*}[t]
	\centering
	\includegraphics[width=0.96\textwidth]{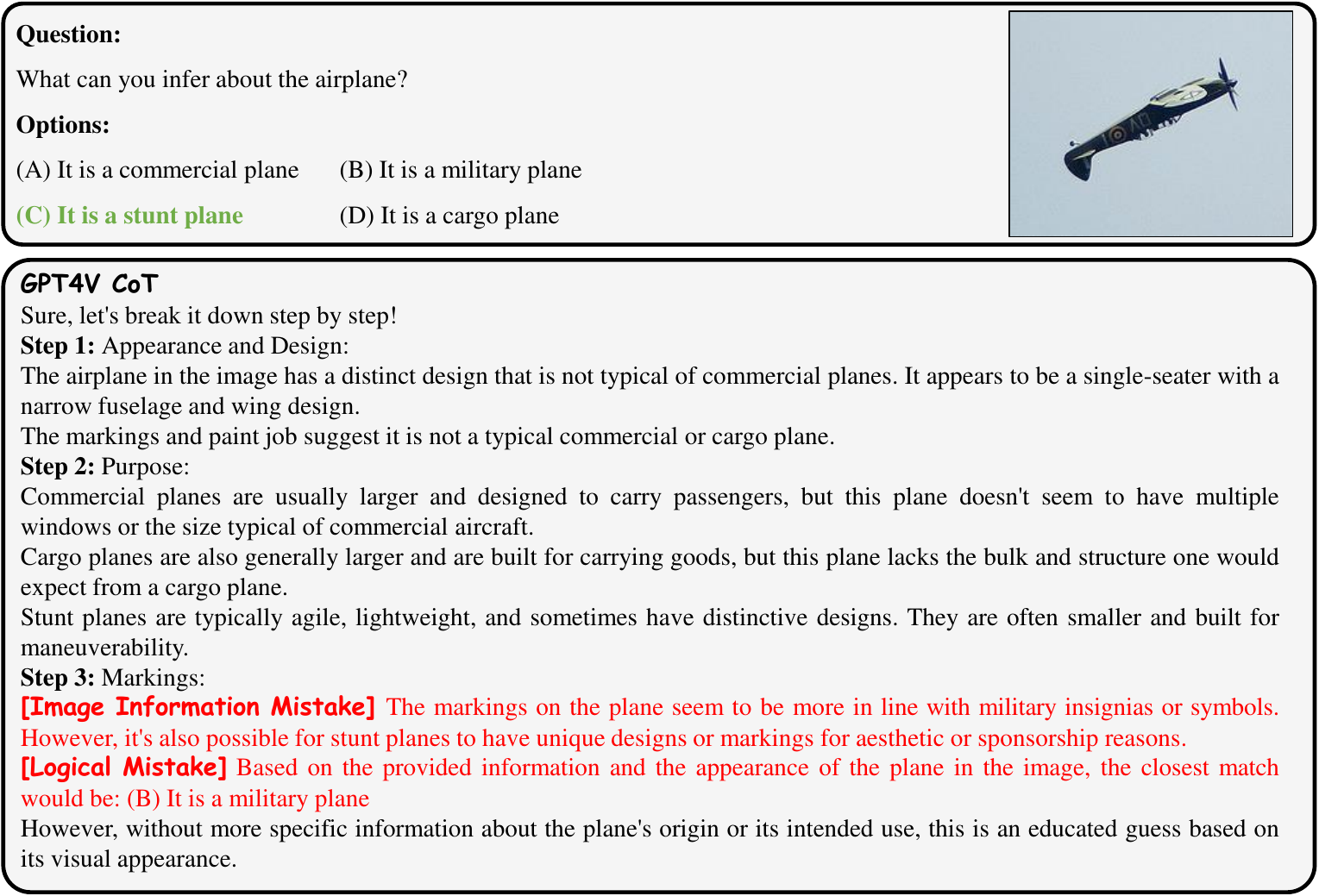}
	\caption{Response from GPT4V with CoT prompting on commonsense domain.}
	\label{fig:error1}
\end{figure*}
\subsubsection{Finetuning Details}
\label{append:finetune}
\paragraph{Model Selection}
Our fine-tuning section incorporates a carefully curated selection of models, which includes a series of traditional Vision-Language Models (VLMs) and Vision Large Language Models (VLLMs). Specifically, VLMs contain MM-CoT~\cite{zhang2023multimodal}, MC-CoT~\cite{tan2023boosting}, and MMR~\cite{wei2023enhancing}. VLLM include {LLaMA-Adapter}~\cite{zhang2023llamaadapter}, {LLaVA-V1.5}~\cite{liu2023improved}, {CogVLM}~\cite{wang2023cogvlm}. This selection was strategically made to encompass a wide range of parameter sizes, architectures, and functionalities. This diversity ensures a comprehensive evaluation of the state-of-the-art in multi-modal capability on finetuning settings.

\paragraph{Experiment Setting}
In the context of finetuning VLLMs, parameter efficient tuning always achieves better performance compared with full-parameter tuning and offers a training cost reduction~\cite{hu2021lora,liu2023improved}. To leverage these benefits, we employ LoRA~\cite{hu2021lora} for parameter-efficient tuning across our experiments. However, for specific cases like the LLaMA-Adapter, we integrate a compact adapter module for fine-tuning, adding minimal additional parameters to the model's architecture.

Our training configurations include a selection of batch sizes from $\{2, 4, 8\}$ and learning rates ranging from $[1e-6, 8e-5]$. We standardize the maximum sequence length to 512 tokens for uniformity across all model trainings. The experiments are conducted on NVIDIA A100 and A800 GPUs to ensure optimal performance and efficiency. For all experiments, model selection is based on the best performance on the development set, which is then validated on the test set for final evaluation.

\subsection{Error Analysis}
\subsubsection{Zero-shot Chain-of-Thought Error Analysis}
\label{sec:zero-cot-case}
\begin{figure*}[h]
	\centering
	\includegraphics[width=0.96\textwidth]{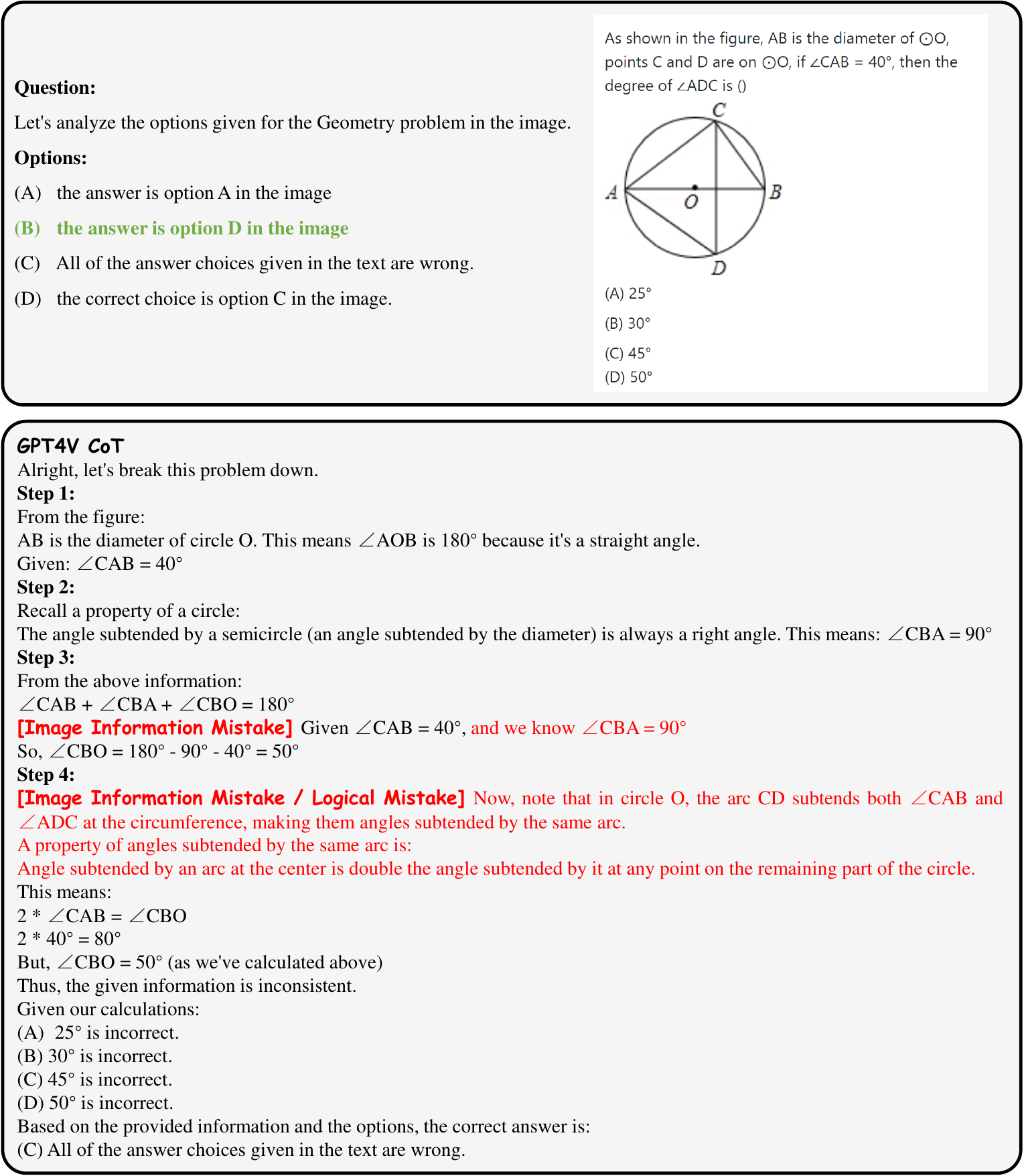}
	\caption{Response from GPT4V with CoT prompting on mathematics domain.}
	\label{fig:error2}
\end{figure*}

\begin{figure*}[h]
	\centering
	\includegraphics[width=0.96\textwidth]{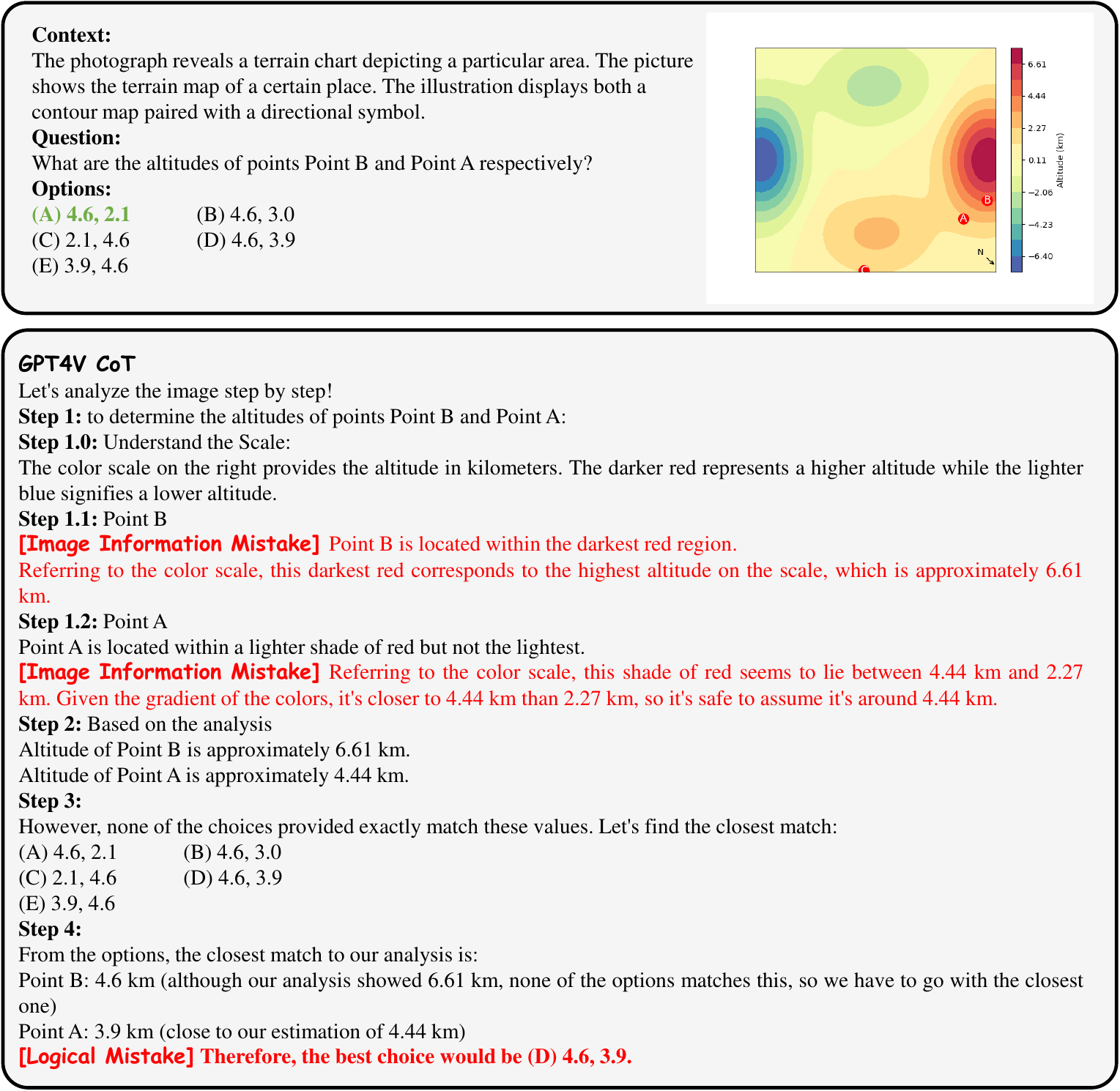}
	\caption{Response from GPT4V with CoT prompting on science domain.}
	\label{fig:error3}
\end{figure*}

In order to further analyze the typical errors in the data set, we analyzed the cases of GPT4V on different domains (as shown in Figure~\ref{fig:error1}, Figure~\ref{fig:error2}, and Figure~\ref{fig:error3}). We can find that all cases have logical errors and visual information interaction errors or deficiencies. This view is also consistent with Section~\ref{sec:analysis} mutual confirmation. Therefore, we believe that the lack of high-quality logical reasoning capabilities and complex multi-modal interactions of large models leads to the failure of multi-step multi-modal reasoning of the model.

\subsubsection{Tool Usage Error Analysis}
\begin{figure*}[h]
	\centering
	\includegraphics[width=0.90\textwidth]{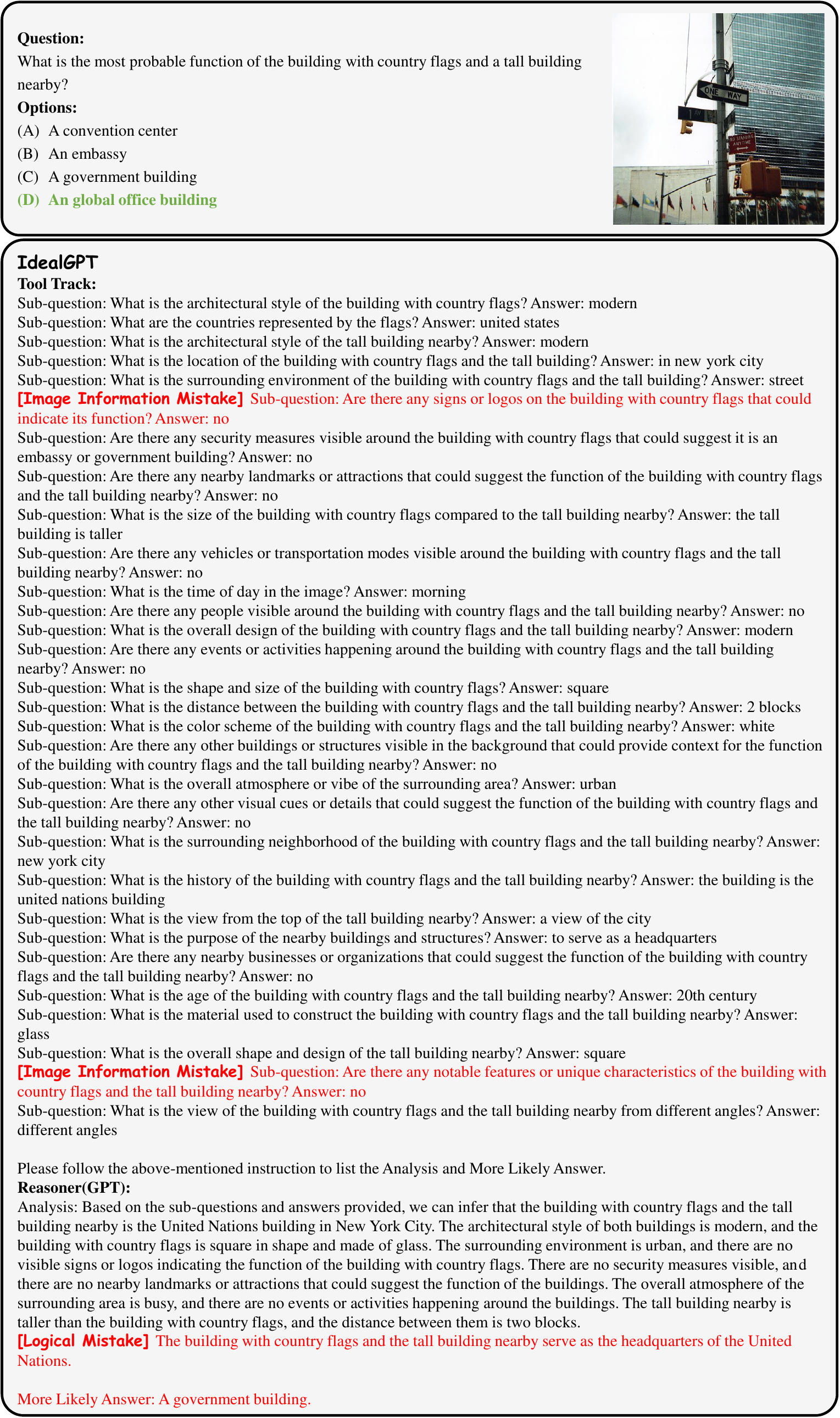}
	\caption{Response from IdealGPT on commonsense domain.}
	\label{fig:error4}
\end{figure*}

\begin{figure*}[h]
	\centering
	\includegraphics[width=0.96\textwidth]{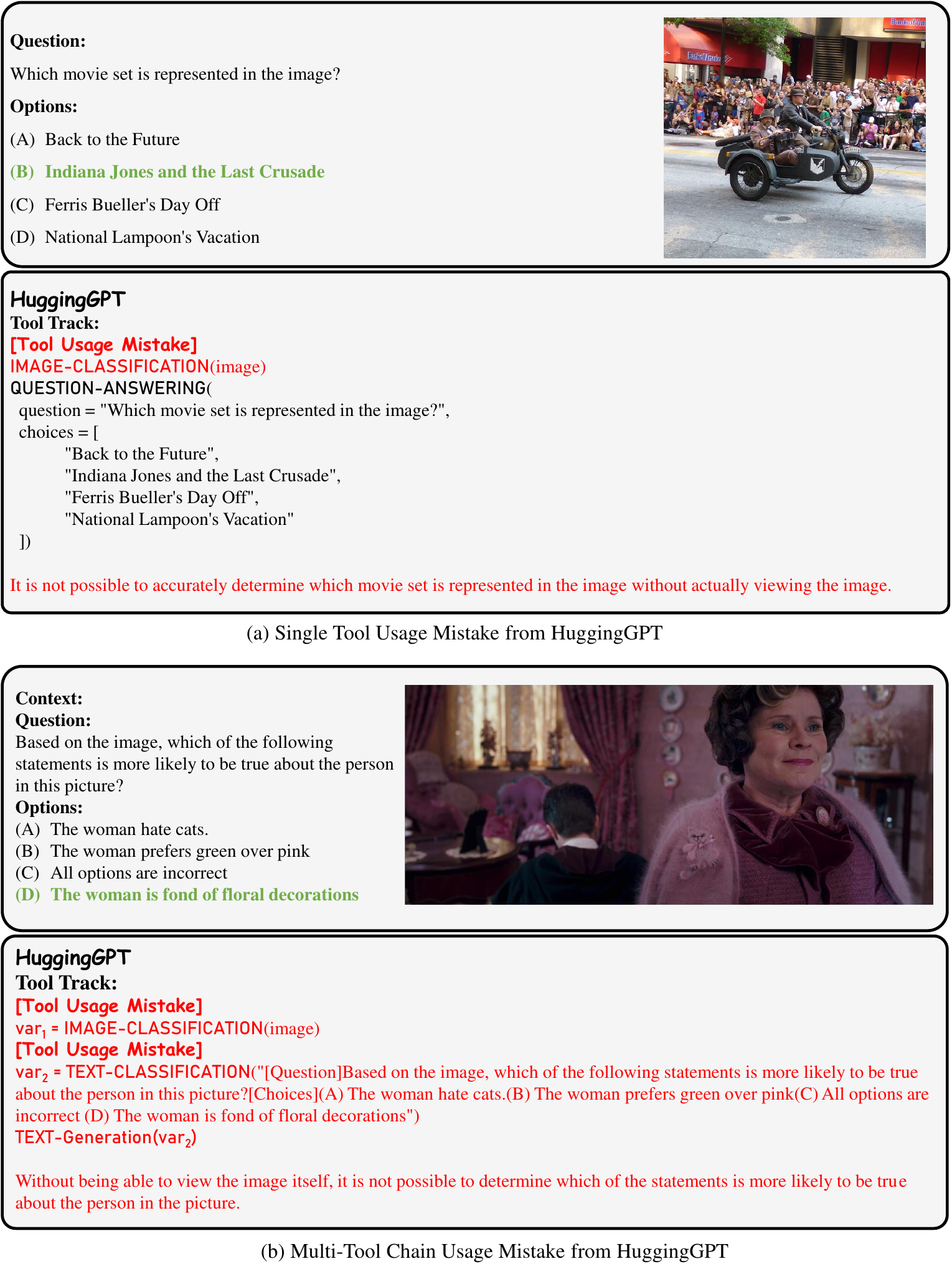}
	\caption{Response from HuggingGPT on commonsense domain.}
	\label{fig:error5}
\end{figure*}

In the context of tool-usage methodologies, an initial challenge emerges from the image information mistake and logical mistake, as highlighted in Appendix~\ref{sec:zero-cot-case}. This limitation becomes particularly problematic in complex scenarios involving multiple tools and steps, as seen in multi-modal task planning. These scenarios demand precise tool selection and sequencing; however, the lack of visual interaction during tool planning leads to frequent errors in tool selection (as shown in Figure~\ref{fig:error5}) and tool-chain redundancy(as shown in Figure~\ref{fig:error4}). Incorrect tool planning or selection can cascade through the process, culminating in complete failure of the intended task (as shown in Figure~\ref{fig:error5}). This issue underscores the need for enhanced model capabilities in processing and integrating visual modalities to accurately navigate multi-step, multi-tool workflows on \datasetname{}.
\label{sec:tool-case}

